%% file: main.tex
\documentclass[]{template}

\usepackage[utf8]{inputenc}             % allow utf-8 input
\usepackage[T1]{fontenc}                % use 8-bit T1 fonts
\usepackage{url}                        % simple URL typesetting
\usepackage{booktabs}                   % professional-quality tables
\usepackage{multirow}
\usepackage{colortbl}
\usepackage{multicol}
\usepackage{amsfonts}                   % blackboard math symbols
\usepackage{nicefrac}                   % compact symbols for 1/2, etc.
\usepackage{microtype}                  % microtypography
\usepackage[dvipsnames]{xcolor}         % colors

\usepackage{latexsym}

\usepackage{graphicx}
\usepackage{float}
\usepackage{subcaption}
\usepackage{wrapfig}
\usepackage{lipsum}

\usepackage{bm}

\usepackage{tabularx} 
\usepackage{ragged2e} 
\newcolumntype{L}{>{\RaggedRight\hangafter=1\hangindent=0em}X}

\usepackage{enumitem}

\usepackage{amsmath}
\usepackage{amssymb}
\usepackage{mathtools}
\usepackage{amsthm}

\usepackage{makecell}
\usepackage{listings}

\setboolean{logo}{true}    % 设为 true 表示显示 logo；设为 false 则不显示

\usepackage[linesnumbered,ruled,vlined]{algorithm2e}

\hypersetup{
    colorlinks=true,
    linkcolor=red,
    citecolor=Cerulean,
    filecolor=magenta,      
    urlcolor=magenta,
}

\usepackage[capitalize,noabbrev]{cleveref}
\crefname{section}{§}{§§}
\Crefname{section}{§}{§§}

\usepackage{calligra}
\DeclareMathAlphabet{\mathcalligra}{T1}{calligra}{m}{n}

\usepackage{pifont}

\theoremstyle{plain}

\theoremstyle{definition}

\theoremstyle{remark}

\renewcommand{\paragraph}[1]{\vspace{1mm}\noindent\textbf{#1}}

\DeclareCaptionLabelFormat{cont}{#1~#2\alph{ContinuedFloat}}
\usepackage[most]{tcolorbox}
\tcbset{
  promptbox/.style={
    top=10pt,
    colback=lightgray!20,
    colframe=Black,
    colbacktitle=NavyBlue,
    enhanced,
    center,
    attach boxed title to top center={yshift=-0.1in,xshift=0.0in},
    boxed title style={boxrule=0pt,colframe=white,},
  }
}
\newtcolorbox{promptbox}[2][]{promptbox, title=#2,#1}
\tcbset{
  takeawaybox/.style={
    top=10pt,
    colback=lightgray!20,
    colframe=Black,
    colbacktitle=BurntOrange,
    enhanced,
    center,
    attach boxed title to top center={yshift=-0.1in,xshift=0.0in},
    boxed title style={boxrule=0pt,colframe=white,},
  }
}
\newtcolorbox{takeawaybox}[2][]{takeawaybox, title=#2,#1}
\tcbset{
  observationbox/.style={
    top=10pt,
    colback=lightgray!20,
    colframe=Black,
    colbacktitle=YellowGreen,
    enhanced,
    center,
    attach boxed title to top center={yshift=-0.1in,xshift=0.0in},
    boxed title style={boxrule=0pt,colframe=white,},
  }
}
\newtcolorbox{observationbox}[2][]{observationbox, title=#2,#1}

\usepackage{xspace}

\newcommand\blfootnote[1]{%
  \begingroup
  \renewcommand\thefootnote{}\footnote{#1}%
  \addtocounter{footnote}{-1}%
  \endgroup
}

\usepackage{CJK}

\definecolor{darkblue}{rgb}{0, 0, 0.5}
\definecolor{myblue}{RGB}{0, 0, 200}
\definecolor{codebg}{gray}{0.95}

\newcolumntype{P}[1]{>{\raggedright\arraybackslash}m{#1}}
\newcolumntype{C}[1]{>{\centering\arraybackslash}m{#1}}
\newcolumntype{L}[1]{>{\raggedright\arraybackslash}m{#1}}

\newcommand{\tinyit}[1]{\textit{\tiny{#1}}}
\newcommand{\scriptbf}[1]{\textbf{\scriptsize{#1}}}

\newcommand{\eg}{\emph{e.g.}\xspace}

\newtcolorbox{mybox}[1]{
    enhanced,
    breakable,
    colback=white,
    colframe=myblue,
    coltitle=white,
    fonttitle=\bfseries\sffamily,
    title={\strut #1},
    rounded corners,
    arc=2mm,
    boxrule=1pt,
    left=4mm, right=4mm, top=4mm, bottom=3mm,
    width=\textwidth,
    boxed title style={
        colback=myblue,
        colframe=myblue,
        boxrule=0pt,
        width=\textwidth,
        sharp corners,
    },
    before upper={\setlength{\parskip}{0.5em}},
}

\title{TREX: Automating LLM Fine-tuning via Agent-Driven Tree-based Exploration}

\author[1]{Zerun Ma$\dagger$}
\author[1]{Guoqiang Wang$\dagger$}
\author[1]{Xinchen Xie$\dagger$}
\author[1,2]{Yicheng Chen}
\author[1,2]{He Du}
\author[1]{Bowen Li}
\author[1]{Yanan Sun}
\author[1]{Wenran Liu}
\author[1]{Kai Chen$\ddagger$}
\author[1]{Yining Li$\ddagger$}

\affil[1]{Shanghai AI Laboratory}
\affil[2]{Fudan University}

\input{sections/0_abstract}

\begin{document}

\blfootnote{$\dagger$ Equal contribution.}
\blfootnote{$\ddagger$ Corresponding authors: Yining Li (liyining@pjlab.org.cn), Kai Chen (chenkai@pjlab.org.cn)}
\blfootnote{$*$ Code and data will be available at \url{https://github.com/trex-project}}

\maketitle

\input{sections/1_introduction}

\input{sections/2_related_work}
\input{sections/3_method}
\input{sections/4_benchmark}
\input{sections/5_experiments}
\input{sections/6_conclusion}

\clearpage
\bibliographystyle{plain}
\bibliography{refs}

%%%%%%%%%%%%%%%%%%%%%%%%%%%%%%%%%%%%%%%%%%%%%%%%%%%%%%%%%%%%

\clearpage
\appendix
\input{sections/appendix}

%%%%%%%%%%%%%%%%%%%%%%%%%%%%%%%%%%%%%%%%%%%%%%%%%%%%%%%%%%%%

% \newpage
% \input{sections/checklists}

\end{document}

%% file: sections/0_abstract.tex
\begin{abstract}
\label{sec:abs}
While Large Language Models (LLMs) have empowered AI research agents to perform isolated scientific tasks, automating complex, real-world workflows, such as LLM training, remains a significant challenge.
In this paper, we introduce TREX, a multi-agent system that automates the entire LLM training life-cycle. By orchestrating collaboration between two core modules—the Researcher and the Executor—the system seamlessly performs requirement analysis, open-domain literature and data research, formulation of training strategies, preparation of data recipes, and model training and evaluation. The multi-round experimental process is modeled as a search tree, enabling the system to efficiently plan exploration paths, reuse historical results, and distill high-level insights from iterative trials.
To evaluate the capability of automated LLM training, we construct FT-Bench, a benchmark comprising 10 tasks derived from real-world scenarios, ranging from optimizing fundamental model capabilities to enhancing performance on domain-specific tasks.
Experimental results demonstrate that the TREX agent consistently optimizes model performance on target tasks.
\end{abstract}

%% file: sections/1_introduction.tex
%%%%%%%%%%%%%%%%%%%%%%%%%%%%%%%%%%%%%%%%%%%%
% Teaser
%%%%%%%%%%%%%%%%%%%%%%%%%%%%%%%%%%%%%%%%%%%
\begin{figure}[h!]
\centering
\includegraphics[width=0.95\textwidth]{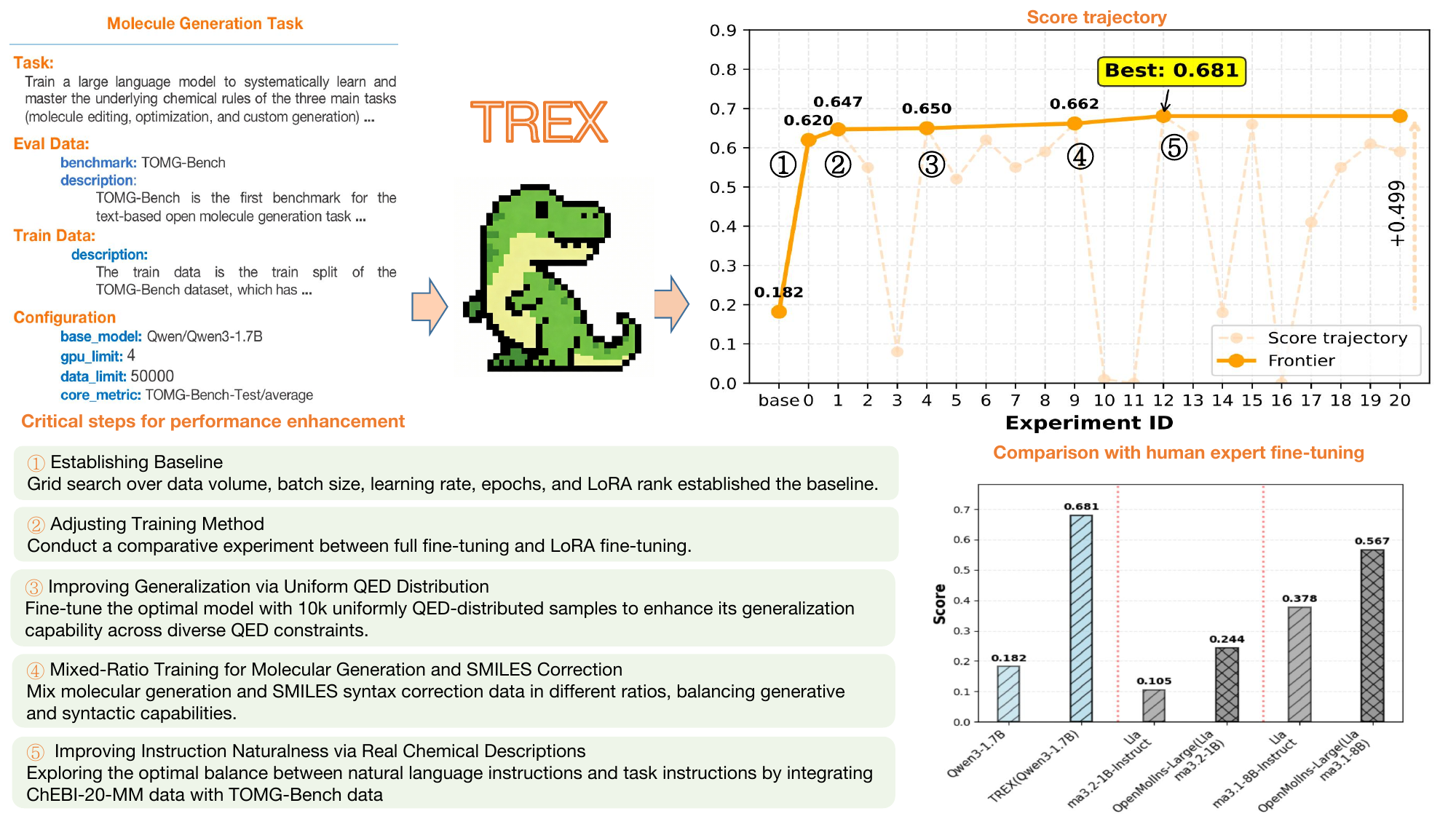}
\captionof{figure}{Illustration of TREX research progress on molecule generation. \textbf{Top Left:} The system takes the task definition as input, including task descriptions, benchmarks, and other information. \textbf{Top Right:} The model performance improves as TREX's automated research progresses. \textbf{Bottom left:} Details of the key steps. \textbf{Bottom right:} Comparison of TREX (based on Qwen3-1.7B) with human-expert fine-tuned models based on Llama-series baselines.}
 \label{fig:teaser}
\end{figure}

\section{Introduction}
\label{sec:intro}
%%%%%%%%%%%%%%%%%%%%%%%%%%%%%%%%%%%%%%%%%%%
% Recent Advances
%%%%%%%%%%%%%%%%%%%%%%%%%%%%%%%%%%%%%%%%%%%
% LLMs become powerful scientific tools
Advancements in artificial intelligence are paving the way for the automation of scientific discovery. Large Language Models (LLMs), equipped with expert-level domain knowledge and long-form generation and comprehension capabilities, have emerged as powerful tools in critical research workflows, including literature review~\cite{zhang2025deep}, hypothesis generation~\cite{gottweis2025towards}, and scientific writing~\cite{lu2024ai}.
% LLM-based agents
Narrowing our scope to AI research, leveraging LLMs' exceptional coding and planning abilities allows recent AI Researcher Agents to autonomously manage the entire experimental lifecycle, from design to execution and refinement. This achievement marks the realization of a fully automated, closed-loop research workflow and yields impressive results in tasks such as machine learning model optimization~\cite{chan2024mle,chi2024sela,liu2025mlmasteraiforaiintegrationexploration} and compute kernel design~\cite{alphaevolve,li2025fm}.

%%%%%%%%%%%%%%%%%%%%%%%%%%%%%%%%%%%%%%%%%%%
% Challenges
%%%%%%%%%%%%%%%%%%%%%%%%%%%%%%%%%%%%%%%%%%%
However, the training of LLMs themselves---one of the most foundational tasks in the current AI landscape---continues to pose significant challenges for current automated approaches.
% Challenge 1: 与代码或文本生成相比，理解和处理训练数据对于 LLM 来说是更大的挑战。
On the one hand, existing research agents typically focus on well-defined optimization targets that can be represented as finite-length text sequences, such as structured parameters~\citep{pmlr-v280-liu25c} or code patches~\citep{novikov2025alphaevolve}. Since these objects can be efficiently encoded and generated by LLMs, they naturally enable an LLM-driven automated workflow that achieves iterative optimization by continuously deriving new proposals from historical experiments. In contrast, designing and optimizing LLM training schemes is a far more open-ended and complex undertaking. Even if we limit the scope to the fine-tuning stage—where the model architecture and initialization weights are fixed—a typical training scheme still requires considering multiple factors, including the training data distribution, algorithms, and hyperparameters. Crucially, while training data has a decisive impact on results, its massive scale makes it impossible to place it directly in the agent's context, posing a substantial barrier to automating scheme design and experiment execution.
% Challenge 2: 方案的验证，模型训练和评测耗时耗资源
On the other hand, unlike tasks such as algorithm design~\cite{imajuku2025ale} or kernel optimization, which allow rapid validation, the training and evaluation of LLMs entail significant time and computational overhead. Consequently, evolutionary frameworks that rely on batch proposal generation and verification become inefficient or even infeasible.
% Challenge 3: Task-grounded, address the needs of real-world scenarios.

%%%%%%%%%%%%%%%%%%%%%%%%%%%%%%%%%%%%%%%%%%%
% Our Contributions
%%%%%%%%%%%%%%%%%%%%%%%%%%%%%%%%%%%%%%%%%%%
% Agent Scaffold
To address these challenges, we introduce TREX, the first automated research agent dedicated to LLM fine-tuning. Based on the given open-ended task objectives, TREX autonomously manages the entire research pipeline by orchestrating collaboration between two core modules: The \textbf{Researcher} interprets the requirement goals to conduct literature reviews and formulate experimental plans. Subsequently, the \textbf{Executor}, a code agent integrated with the GPU clusters, implements the experimental plans to construct data and perform model training and evaluation, closing the loop by feeding empirical results back to the Researcher for analysis.

% Tree-search
Human researchers typically optimize model training strategies through an iterative process of experimentation, exploring novel approaches while retaining those empirically validated as effective. Inspired by this fundamental research paradigm, we designed a Monte Carlo Tree Search (MCTS)-based approach to automate the exploration of experimental schemes. By selecting and optimizing the most promising paths from historical experiments, this strategy reliably generates high-quality experimental plans without requiring extensive rollouts. Additionally, we equip the agent with a comprehensive suite of tools to support academic web search, experiment history tracking, and cluster task orchestration. A key component of this toolkit is the AI Data Processing (AIDP) library, which provides high-performance data-processing primitives tailored to LLM training scenarios. The agent can compose complex training data pipelines with AIDP tools, ensuring more reliable and efficient model training. Furthermore, to accelerate the iteration loop, we incorporated a fine-grained analysis mechanism into the evaluation phases. By synthesizing task performance metrics, cross-model comparisons, and bad-case analysis, the agent maximizes the informative feedback from each experiment to guide subsequent optimization.

% Benchmark
To evaluate the efficacy of TREX, we introduce FT-Bench, a benchmark designed to assess automated scientific research systems specifically on LLM fine-tuning tasks. FT-Bench comprises 10 evaluation datasets from academic papers and competitions, covering diverse task types ranging from general capability enhancement to vertical domain adaptation. This Benchmark is designed to measure the system's capacity to improve LLM performance on these datasets through fine-tuning.

% Summary
Experimental results demonstrate that our proposed TREX system can continuously optimize LLM fine-tuning strategies and data under limited resource and time budgets, effectively enhancing model performance on the evaluation tasks. Notably, on several tasks, TREX achieves performance gains that surpass those of expert-designed fine-tuning recipes. While powering TREX with SOTA proprietary LLMs like gemini3-pro and claude4.5-sonnet yields the best results, the system remains functional and effective when driven by open-source models Qwen3-Next-80B~\cite{qwen3technicalreport}. Additionally, ablation studies validate the effectiveness of our proposed tree-based exploration strategy and fine-grained evaluation mechanism.

%% file: sections/2_related_work.tex
\section{Related Work}
\label{sec:related_work}
% 可参考
% - deprecated-1
% - https://aicarrier.feishu.cn/wiki/TLv8wWRAqisdCGkVek9conzEnsc

%%%%%%%%%%%%%%%%%%%%%%%%%%%%%%%%%%%
\subsection{AI-Augmented Research}
% AI technologies that facilitate specific research tasks and serve as tools or assistants for human researchers.
As large language models (LLMs) and LLM-based agents advance rapidly, AI has evolved from a rudimentary tool to an indispensable collaborative partner in the research process.

%%%%%%%%%%%%%%%%%%%%%%%%%%%%%%%%%%%
\noindent\textbf{Literature Discovery.} % Deep-Research & RAG
Literature discovery is essential yet labor-intensive in traditional academic research. AI tools like DeepResearch~\cite{zhang2025deep} and retrieval-augmented generation (RAG)~\cite{lewis2020rag,karpukhin2020dense} significantly reduce the time spent on literature retrieval and summarization, thereby boosting the efficiency of scholarly investigation.

\noindent\textbf{Paper Writing and Review.} % AI for paper writing and review. Should mention the relevant academic ethical risks.
LLMs excel at writing and are increasingly being used to assist in academic writing. Current AI-Agent systems \cite{lu2024aiscientist,aiscientist_v2} have demonstrated the ability to autonomously generate complete, peer-reviewed academic papers. Some conference reviewers also use AI as an auxiliary evaluation tool. Although there are academic ethics concerns, the convenience and efficiency of AI for researchers are undeniable.

\noindent\textbf{Idea Generation and Implementation.} % AI that generates idea/hypothesis and implements code, [with human collaborations]
AI agents have inspired broader AI applications in scientific discovery. Systems like Dolphin\cite{yuan2025dolphin}, InternAgent\cite{team2025novelseek}, AI-Scientist v1\&v2\cite{lu2024aiscientist,aiscientist_v2}, AI-Researcher\cite{tang2025ai} have proven feasible for AI to autonomously generate research ideas and implement experiments. Although human oversight is still needed, these advances mark a key step toward automating AI-driven scientific research.

%%%%%%%%%%%%%%%%%%%%%%%%%%%%%%%%%%%
\subsection{Autonomous Research Agents}
%%%%%%%%%%%%%%%%%%%%%%%%%%%%%%%%%%%

\noindent\textbf{End-to-End AI Researcher.} % AI Scientist, AI Researcher,... 局限性：缺少任务导向，偏向于验证自动化科研这一概念，但缺少对效果的系统性量化评估
Previous studies~\cite{lu2024aiscientist,aiscientist_v2,tang2025ai} have achieved notable advancements in automated scientific research. Specifically, AI Scientist v1\&v2\cite{lu2024aiscientist,aiscientist_v2} encompasses core research workflows, including idea generation, experimental implementation, and academic paper composition. In contrast, AI Researcher~\cite{tang2025ai} further integrates additional essential functions beyond these, such as literature review, algorithm verification and optimization, and result analysis. These studies have essentially implemented the workflow of LLM-based automated scientific research; however, their primary contribution lies in verifying the feasibility of end-to-end AI-driven research, without incorporating clear task orientation or conducting a quantitative evaluation of the research system's performance.

\noindent\textbf{Evolutionary Search Agent.} % AlphaEvolve, FM Agent, SELA, AIDE，ML-Master, DeepScientist ... 局限性：依赖批量生成方案和验证 (batch generation and verification)，对大模型训练这种高计算开销的任务效率低下
A series of studies~\cite{alphaevolve,li2025fm,chi2024sela,aide2025,liu2025mlmasteraiforaiintegrationexploration,weng2025deepscientist} have incorporated methodologies such as evolutionary algorithms and tree search into automated scientific discovery. These approaches offer substantial support for generating and implementing diverse research ideas. Nevertheless, these methods are highly dependent on large-scale method sampling, making them practically unable to complete tasks with high computational overhead—specifically, LLM fine-tuning.

\noindent\textbf{Benchmarking Automated AI Research.} % MLE-Benchmark, RE-Benchmark, ... 局限性：没有针对 LLM 训练的（系统性）评测集
Benchmarks such as MLE-bench \cite{chan2024mle} and RE-bench \cite{wijk2024re} are widely used for evaluating automated AI research systems. MLE-bench assesses AI agents' performance on machine learning tasks, while RE-bench evaluates their capabilities across key aspects, including code synthesis and experimental implementation. Existing benchmarks, including IdeaBench \cite{guo2025ideabench}, LiveIdeaBench \cite{ruan2024liveideabench}, and DiscoveryWorld \cite{jansen2024discoveryworld}, effectively evaluate AI-agents in relevant scenarios but do not provide a systematic evaluation dataset tailored for LLM training—a gap to be addressed.
% The evaluation of AI research agents remains fragmented, given that existing frameworks—including AI-Scientist-v2 \cite{yamada2025ai} and AgentLab \cite{schmidgall2025agentrxiv}—rely on ad-hoc datasets and proprietary methodologies, thus hindering the development of a standardized benchmarking paradigm. Current evaluative efforts encompass three core dimensions: epistemic novelty and credibility (assessed via IdeaBench \cite{guo2025ideabench} and LiveIdeaBench \cite{ruan2024liveideabench}, which focus on hypothesis generation and scientific conceptualization), engineering rigor and efficiency (quantified by suites such as MLE-bench \cite{chan2024mle}, SWE-bench \cite{jimenez2023swe}, and RE-bench \cite{wijk2024re} to measure proficiency in code synthesis and experimental implementation), and end-to-end orchestration (evaluated through frameworks like \cite{chen2024scienceagentbench} and DiscoveryWorld \cite{jansen2024discoveryworld} for holistic completion of complex scientific workflows). Despite the coverage of these three core evaluative dimensions, critical deficits persist: a critical gap lies in the absence of a (systematic) evaluation dataset tailored for LLM training.

%%%%%%%%%%%%%%%%%%%%%%%%%%%%%%%%%%%
\subsection{Automated Model Training}
%%%%%%%%%%%%%%%%%%%%%%%%%%%%%%%%%%%

\noindent\textbf{AutoML.} % Technologies like NAS, LLM designed architecture/hyperparameters/reward（可参考 Towards Execution-Grounded Automated AI Research 论文 Related work 部分）局限性：在限定条件下优化固定的单一目标，而不是在开放环境下直接设计和优化实验方案
Traditional autonomous machine learning~\cite{le2020scaling,erickson2020autogluontabularrobustaccurateautoml} primarily focuses on automating model selection and hyperparameter configuration.
More recently, a line of work~\cite{tang2024autogluonmultimodalautommsuperchargingmultimodal} has leveraged LLMs to generate architecture variants~\cite{cheng2025languagemodelinglanguagemodels} and synthesize post-training objectives~\cite{lu2024discoveringpreferenceoptimizationalgorithms}.
These methods remain constrained by predefined search spaces or focus on optimizing isolated components. In contrast, our work explores a more open-ended setting, directly automating the entire LLM training lifecycle.

\noindent\textbf{AI for Data Construction.} % 1）利用 LLM 构造训练数据；2）自动化数据 Agent... 局限性：更多作为工具，依赖人工设计 data pipeline
Recent studies extensively utilize LLMs for data synthesis~\cite{mitra2024agentinstructgenerativeteachingagentic}, evolutionary refinement~\cite{xu2025wizardlmempoweringlargepretrained}, and quality filtering~\cite{liu2024what}.
To facilitate these tasks, dedicated frameworks~\cite{djv1,liang2025dataflow} have been developed to ensure reproducible and scalable data engineering.
However, these approaches typically employ LLMs as discrete tools within predetermined protocols. In contrast, we propose a holistic and fully automatic paradigm that integrates data preparation into an autonomous loop.

%% file: sections/3_method.tex
\section{TREX}
\label{sec:method}

%%%%%%%%%%%%%%%%%%%%%%%%%%%%%%%%%%%%%%%%%%%%%%%%%%%%%
\subsection{Overview}
\begin{figure*}[!ht]
    \centering
    \vspace{-5pt}
    \includegraphics[width=0.9\textwidth, trim=0cm 2.5cm 0cm 0cm, clip]{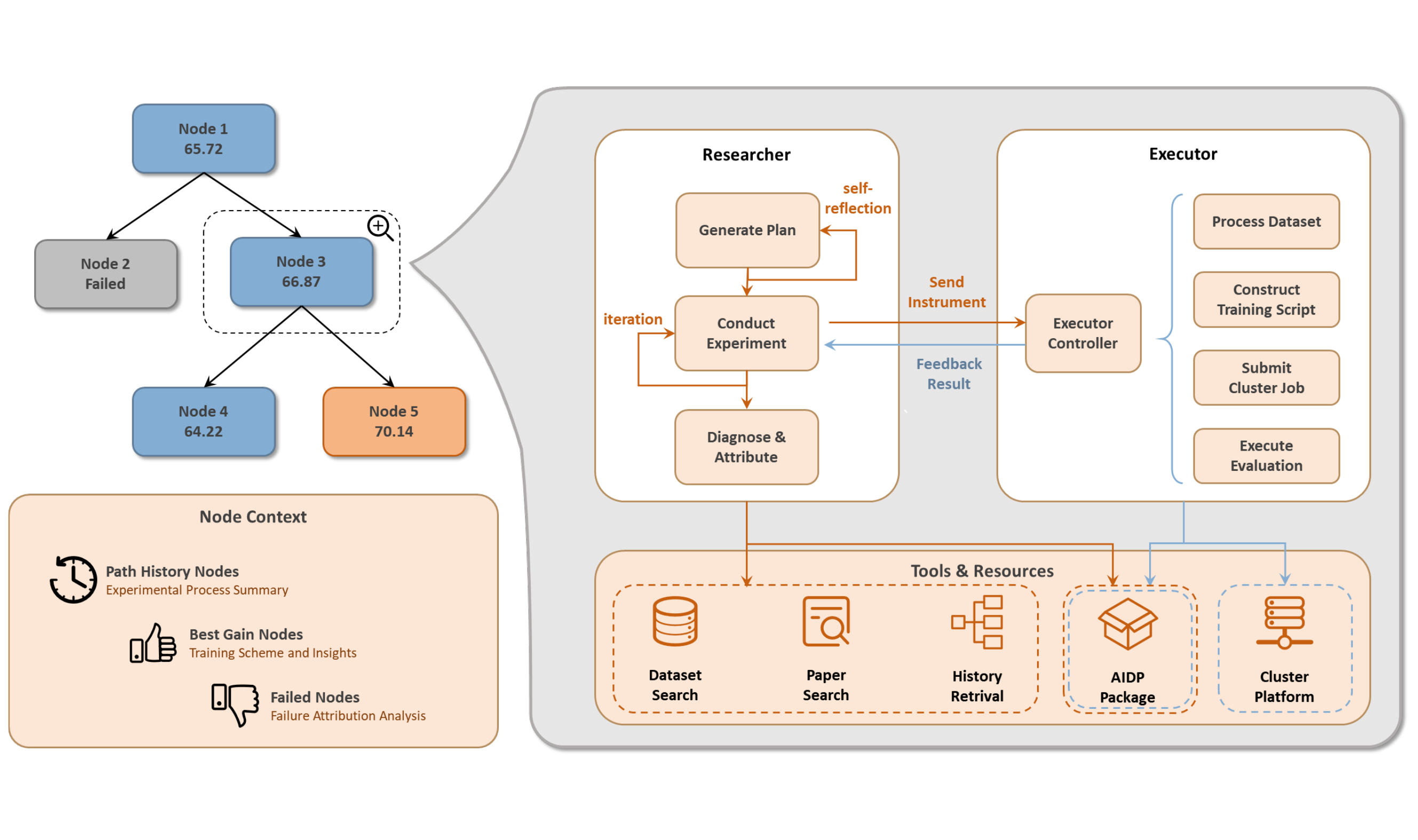}
    \caption{The TREX framework. The system performs task-driven model optimization through iterative experimentation. The experimental trajectory is modeled as a tree, with node expansion guided by MCTS policy. Each iteration is driven by dual-agent collaboration: The \textit{Researcher} designs the experimental plan and delegates the specific steps to the \textit{Executor}. The \textit{Executor} performs data processing, model training, and evaluation on a GPU cluster and returns empirical results to the \textit{Researcher} for diagnostics. To support this workflow, the agents are equipped with a tailored toolset to access external knowledge and resources, and leverage condensed memory contexts to preserve the continuity and efficiency of the iterative process.}
    \label{fig:framework}
    \vspace{-2pt}
\end{figure*}

%%%%%%%%%%%%%%%%%%%%%%%%%%%%%%%%%%%%%%%%%%%%%%%%%%%%%
Figure~\ref{fig:framework} illustrates the multi-agent design and the entire workflow of the proposed TREX framework. For an LLM fine-tuning task, TREX employs a dual-loop workflow to automatically conduct experimental iterations and optimize solutions. Within the inner loop, two agents, namely Researcher and Executor, collaborate to complete a single-round experiment (Sec.~\ref {sec:agent}). Specifically, the Researcher first formulates an experimental plan for the current iteration based on task objectives and historical records. Subsequently, the Researcher initiates multi-round communication with the Executor, instructing the Executor to execute the plan step by step. The Executor is deployed on a GPU cluster and translates received instructions into executable code. Upon completing model training and evaluation, the Researcher analyzes the results and concludes the current experimental cycle.

In the outer loop, the multi-round experimental process is modeled as a tree, where each new round corresponds to creating a new node. We designed an MCTS-based strategy to expand the set of experiment nodes, striking an efficient balance between exploiting high-performing existing solutions and exploring novel ones.

%%%%%%%%%%%%%%%%%%%%%%%%%%%%%%%%%%%%%%%%%%%%%%%%%%%%%
\subsection{Agent Framework}
\label{sec:agent}
%%%%%%%%%%%%%%%%%%%%%%%%%%%%%%%%%%%%%%%%%%%%%%%%%%%%%

\noindent\textbf{Researcher.}
The Researcher Agent serves as the core module responsible for experimental design and result analysis. In each experimental iteration, the Researcher employs a coarse-to-fine approach to formulate the experimental scheme. Specifically, it first determines a high-level improvement strategy based on contextual information, such as augmenting training data or exploring alternative training algorithms. Subsequently, the Researcher refines this strategy into a concrete experimental plan that typically includes specific objectives, data processing procedures, and training configurations. To support this workflow, the Researcher can use academic search engines to retrieve and digest relevant literature from arXiv and locate accessible open-source datasets via Hugging Face. Each experimental plan typically comprises 3–5 parallel training configurations to investigate the impact of variables such as data mixing ratios and hyperparameters. This approach not only enhances iteration efficiency but also maximizes the utilization of cluster computing resources. 

\noindent\textbf{Executor.}
We implemented the Executor agent based on OpenHands~\cite{wang2024openhands} to leverage its robust capabilities in task planning and code implementation. Furthermore, we encapsulated the cluster task-scheduling APIs as tools and integrated them into the Executor, enabling it to use and monitor cluster resources. Meanwhile, the Executor agent's built-in sandboxing system ensures the experimental environment remains isolated, preventing file read/write operations from posing risks to cluster data.

\noindent\textbf{AIDP Library.}
To enhance the system's capability in curating model training data, we introduce AIDP (AI Data Processor), a modular library tailored for developing LLM data pipelines.
AIDP provides a suite of high-performance, reliable, and deterministic operators built upon the HuggingFace Datasets ecosystem, ensuring experimental reproducibility and parallel processing to minimize overhead in computationally intensive tasks.
Furthermore, AIDP strikes a strategic balance between granularity and abstraction: its primitives are semantically clear for precise interpretation and invocation, yet sufficiently modular to enable agent systems such as TREX to freely reconfigure and orchestrate complex pipelines. A detailed introduction of AIDP can be found in Appendix~\ref{sec:appendix_aidp}

\noindent\textbf{Experiment Diagnostics and Attribution.}
Evaluation feedback is pivotal for iterative improvement of experimental schemes. While benchmark scores serve as direct optimization targets, they often lack sufficient granularity to effectively guide refinements to training schemes. Given the substantial computational costs of LLM training, it is imperative to extract rich feedback from each iteration to accelerate exploration toward an optimal scheme. To this end, we employ a holistic diagnostic analysis strategy. For each experiment, the agent examines failure cases within the validation set and conducts an attribution analysis. Furthermore, the agent performs a comparative analysis of benchmark metrics and bad-case attributions across the current and historical experiments to identify the determinants of performance shifts. These insights are synthesized into an evaluation report and stored in the context memory to inform subsequent experimental designs.

%%%%%%%%%%%%%%%%%%%%%%%%%%%%%%%%%%%%%%%%%%%%%%%%%%%%%
\subsection{Iterative Exploration for Training Schemes}
\label{sec:tree_search}
%%%%%%%%%%%%%%%%%%%%%%%%%%%%%%%%%%%%%%%%%%%%%%%%%%%%%

\noindent\textbf{MCTS.}
Prior works on AutoML~\cite{chi2024sela,liu2025mlmasteraiforaiintegrationexploration} have employed MCTS~\cite{coulom2006efficient} to navigate vast solution spaces, leveraging its ability to balance exploration of novel solutions with exploitation of high-performing existing ones. Drawing inspiration from this paradigm, we use MCTS to drive the iterative exploration of the space of LLM training schemes. Specifically, this process starts at a root node, where the agent conducts an initial round of experiments to establish a baseline, including constructing a functional training dataset and performing a grid search over training hyperparameters. In subsequent iterations, we select the node with the highest Upper Confidence Bound for Trees (UCT) for refinement. A new training scheme is then generated and empirically verified by the agent, after which the UCT scores of nodes throughout the search tree are updated according to the following definition:

\begin{equation}
\label{eqa:uct}
\mathrm{UCT}(v)=\frac{Q_v}{N_v} + c\cdot\sqrt{\frac{\ln{N_{P(v)}}}{N_v}}
\end{equation}

Where $N_v$ and $N_{P(v)}$ denote the total visit counts of the experiment node $v$ and its parent node, respectively. $Q_v$ represents the accumulated reward of node $v$. In our experiments, the reward for each experiment is defined as the normalized value of the primary evaluation metric for the task (see Table~\ref{tab:benchmarks}). To improve exploration efficiency, the agent typically trains a batch of models in parallel with slightly varying configurations (e.g., hyperparameters or data ratios) and uses the best-performing model to compute the reward.

\noindent\textbf{Memory Context Management.}
The agent iteratively designs training schemes using historical records. To prevent accumulating history from causing redundancy or exceeding the LLM's context window, we introduce a memory management strategy. The accessible context for the current node $v$ is constructed as follows:

\begin{equation}
\label{eqa:memory}
\mathrm{MC}(v)=Condense(\mathcal{P}(v),\mathcal{S}(v),\mathcal{C}(Tr))
\end{equation}

The memory context integrates experimental history from three components: $\mathcal{P}(v)$ represents the trajectory from the root to node $v$, capturing the lineage of improvements preceding the current experiment; $\mathcal{S}(v)$ denotes the sibling nodes of $v$, serving to enhance scheme diversity by preventing homogeneous attempts under the same parent; $\mathcal{C}(Tr)$ refers to critical nodes within the current experiment tree $Tr$ that yields significant performance gains or failures, enabling global sharing of key empirical insights. This historical information is condensed into a memory context to facilitate the design and analysis of the current experiment.

%% file: sections/4_benchmark.tex
\section{FT-Bench}
\label{sec:ftbench}

The inherently open-ended and complex nature of AI-driven research presents significant challenges for evaluation. Existing benchmarks predominantly focus on dimensions such as code generation~\cite{jimenez2023swe,chen2024scienceagentbench} and machine learning engineering~\cite{huang2023mlagentbench,chan2024mle}, leaving a substantial gap when applied to frontier LLM training tasks. The frontier LLM training task represents a comprehensive systems engineering effort that integrates research on training algorithms, data engineering, and repository-level code comprehension. To address this gap, we introduce FT-Bench to systematically evaluate agents' capabilities on automating LLM fine-tuning.

%%%%%%%%%%%%%%%%%%%%%%%%%%%%%%%%%%%%%%%%%%%
\subsection{Benchmark Construction}
%%%%%%%%%%%%%%%%%%%%%%%%%%%%%%%%%%%%%%%%%%%

% Task selection
Our benchmark design is guided by two primary principles. First, the evaluation tasks should originate from real-world research or industrial applications to faithfully benchmark the agent's proficiency in actual LLM fine-tuning. Second, the computational and data overhead must be controllable, since excessively compute- or data-heavy tasks would severely bottleneck multi-round experimental iterations. Motivated by these considerations, we ultimately selected 10 distinct LLM fine-tuning tasks to construct FT-Bench~\ref{tab:benchmarks}. 
These tasks encompass common LLM fine-tuning scenarios, ranging from augmenting domain-specific knowledge and proficiencies~\cite{yim2023aci,tang2024autogluonmultimodalautommsuperchargingmultimodal,xu2025omebench,baker2016automatic,chen2024scienceagentbench,eastmoney2023openfindata,fei2024lawbench} to enhancing generalized capabilities, such as tool calling~\cite{wang2024gta} and logical reasoning~\cite{aiscientist_v2,quan2024econlogicqa}.
For each task, we provide the agent with a structured description that includes the task objective, evaluation data and metrics, training data (if applicable), and experimental configurations. Regarding the training data, while an initial dataset is provided for certain tasks, the agent is neither obligated to use it nor restricted to it exclusively. Instead, the agent is granted autonomy to search for, synthesize, and curate training data. For tasks lacking initial training data, the agent must build the training set from scratch. Figure~\ref{fig:teaser} illustrates a sample task description, with further details provided in the Appendix~\ref {sec:appendix_benchmark_detail}.

\begin{table*}[!ht]
\begin{center}
\caption{LLM fine-tuning tasks in FT-Bench. Detailed task definitions are provided in Appendix~\ref{sec:appendix_benchmark_detail}}
\label{tab:benchmarks}
\footnotesize
\renewcommand{\arraystretch}{1.5}
\begin{tabular}{L{0.18\textwidth} L{0.75\textwidth}}
\toprule
\textbf{Task} & \textbf{Description} \\
\midrule
ACI-Bench~\citep{yim2023aci} & \textbf{Clinical Notes Generation}: generate structured visit notes from doctor-patient dialogues \\
% \midrule
TOMG-Bench~\citep{li2024tomg} & \textbf{Molecule generation}: edit, optimize, and generate molecules following textual instructions \\
% \midrule
oMeBench~\citep{xu2025omebench}  & \textbf{Chemical Mechanism Reasoning}: predict stepwise intermediates and electron-transfer steps for organic reactions \\
% \midrule
HoC~\citep{baker2016automatic}& \textbf{Cancer Hallmark Classification}: assign cancer hallmark labels to biomedical publication abstracts \\
% \midrule
CS-Bench~\citep{song2024cs} & \textbf{Computer Science Proficiency Evaluation}: answer knowledge and reasoning questions across 26 subfields of computer science \\
% \midrule
OpenFinData~\citep{eastmoney2023openfindata} & \textbf{Financial Question Answering}: solve real-world financial tasks including numerical calculation, knowledge retrieval, and chart understanding \\
% \midrule
SST-2~\citep{socher2013recursive} & \textbf{Sentiment Classification}: predict the binary sentiment polarity of movie review sentences \\
% \midrule
EconlogicQA~\citep{quan2024econlogicqa} & \textbf{Economic Sequential Reasoning}: rank multiple interconnected events into the correct logical order within economic scenarios \\
% \midrule
GTA~\citep{wang2024gta} & \textbf{Agentic Tool Use}: complete implicit-tool-use tasks grounded in real multimodal inputs through multi-step planning and execution \\
% \midrule
LawBench~\citep{fei2024lawbench} & \textbf{Legal Knowledge Evaluation}: perform 20 Chinese legal tasks spanning memorization, comprehension, and application of legal knowledge \\
\bottomrule
\end{tabular}
\end{center}
\end{table*}

%%%%%%%%%%%%%%%%%%%%%%%%%%%%%%%%%%%%%%%%%%%
\subsection{Comparison with Existing Benchmarks}
%%%%%%%%%%%%%%%%%%%%%%%%%%%%%%%%%%%%%%%%%%%

Table~\ref{tab:benchmark-comparison-en} presents a comparison between FT-Bench and relevant existing benchmarks. Current lines of work on research agent evaluation encompass a broad spectrum of tasks, including ML model optimization~\cite{huang2023mlagentbench}, Kaggle-style engineering challenges~\cite{chan2024mle}, AI R\&D~\cite{wijk2024re,nathani2025mlgym}, and scientific code implementation~\cite{chen2024scienceagentbench}. Despite their breadth, these benchmarks exhibit common limitations: they either evaluate agents on isolated sub-tasks within constrained environments or restrict the scope to traditional ML paradigms, thereby failing to capture the unique challenges of modern LLM training, such as instruction formatting, domain-specific evaluation, and the orchestration of the full training lifecycle. FT-Bench addresses both limitations by providing the first benchmark dedicated entirely to end-to-end LLM fine-tuning and requiring agents to autonomously navigate the complete fine-tuning pipeline in an open environment.

\begin{table*}[!ht]
\begin{center}
\caption{Comparison of benchmarks for autonomous research systems.}
\label{tab:benchmark-comparison-en}
\footnotesize
\renewcommand{\arraystretch}{1.2}
\begin{tabular}{L{0.2\textwidth} L{0.08\textwidth} L{0.62\textwidth}}
\toprule
\textbf{Benchmark} & \makecell[c]{\hspace{-20pt}\textbf{\#Tasks} \\ \hspace{-20pt}
\scriptbf{(LLM-FT~/~Total)}}  & \scriptbf{Evaluation Focus} \\
\midrule
MLAgentBench~\citep{huang2023mlagentbench} & 0/13 & End-to-end machine learning experimentation with iterative model development and code refinement. \\
ScienceAgentBench~\citep{chen2024scienceagentbench} & 0/102 & Fine-grained scientific workflow tasks derived from real research across multiple disciplines. \\
RE-Bench~\citep{wijk2024re} & 1/7 & Open-ended machine learning research engineering tasks with direct comparison to human experts. \\
MLEBench~\citep{chan2024mle} & 2/75 & Real-world machine learning engineering tasks from Kaggle competitions with human leaderboard baselines. \\
MLGymBench~\citep{nathani2025mlgym} & 0/13 & AI research tasks spanning multiple domains, with a Gym-style interface and supporting RL-based agent training.\\
\midrule
\textbf{FT-Bench (ours)} & 10/10 & End-to-end LLM fine-tuning tasks derived from real-world research and application scenarios.\\
\bottomrule
\end{tabular}
\end{center}
\end{table*}

%% file: sections/5_experiments.tex
\section{Expriment}
\label{sec:experiemnts}

%%%%%%%%%%%%%%%%%%%%%%%%%%%%%%%%%%%%%%%%%%%%%%%%%%%%%
\subsection{Experimental Setup}
\label{sec:exp_setup}
%%%%%%%%%%%%%%%%%%%%%%%%%%%%%%%%%%%%%%%%%%%%%%%%%%%%%

\noindent\textbf{Tasks.} We evaluate the proposed TREX framework across 10 tasks from FT-Bench. Each task is fed to the Researcher via a structured template, with the input comprising the task description, evaluation protocol and entry, raw training data (optional, available only for specific tasks), and experimental configurations (see Figure 1 for an example). To ensure time and computational efficiency, we uniformly employ the lightweight model Qwen3-1.7B~\cite{qwen3technicalreport} as the base model for autonomous fine-tuning and cap the maximum number of training samples at 50,000 per experiment.

\noindent\textbf{System Settings.} For the Researcher agent, we verify its capability of conducting autonomous, multi-round experimental design across varying backend LLMs, including an open-source model (Qwen3-Next-80B-Thinking~\cite{qwen3technicalreport}) and a proprietary model (Gemini 3 Pro~\cite{googledeepmind2025gemini3pro}). For the Executor agent, we adopt Claude 4.5 Sonnet, following OpenHands' recommendations, to maximize the robustness of experiment execution.

\noindent\textbf{Evaluation.} We allow TREX to explore a maximum of 20 experimental iterations per task, selecting the best-performing round as the final result. Task-specific evaluation metrics and protocols are detailed in Table~\ref{tab:exp_results_1} and Appendix~\ref{sec:appendix_benchmark_detail}. Given the inherent variance in metric scales and optimization difficulties across different tasks, we introduce a normalized relative performance gain for fair comparison:

\begin{equation}
\label{eqa:rel_gain}
G_T=\frac{\mathcal{E}_T(M_{FT}) - \mathcal{E}_T(M_{Base})}{\mathcal{E}_T(M_{Ref}) - \mathcal{E}_T(M_{Base})}
\end{equation}

Where the $\mathcal{E}_T$ is the task-specific evaluation function, while $M_{Base}$ and $M_{FT}$ represent the models before and after autonomous fine-tuning, respectively. $M_{Ref}$ is a superior reference model, and the performance gap between it and the base model provides a baseline for normalizing and quantifying the relative gains achieved. We use Qwen3-235B-2507~\cite{qwen3technicalreport} as $M_{Ref}$ in experiments.

%%%%%%%%%%%%%%%%%%%%%%%%%%%%%%%%%%%%%%%%%%%%%%%%%%%%%
\subsection{Main results}
%%%%%%%%%%%%%%%%%%%%%%%%%%%%%%%%%%%%%%%%%%%%%%%%%%%%%

Table~\ref{tab:exp_results_1} summarizes the performance of TREX across various FT-Bench tasks. The experimental results indicate that TREX consistently enhances the base model's performance across all evaluated tasks through iterative exploration of the fine-tuning scheme. 
Furthermore, configuring the Researcher agent with Gemini 3 Pro as the reasoning backend yields superior performance compared to the Qwen3-Next-80B-Thinking backend on the vast majority of tasks. This demonstrates that the intrinsic reasoning capability of the underlying LLM directly impacts the overall efficacy of the TREX system.
Empirical observations reveal that TREX's optimization efficacy varies considerably across tasks. When an initial training set is provided (\eg, ACI-Bench and TOMG-Bench), it facilitates the system to more easily establish a strong baseline and efficiently iterate over data and training schemes. Conversely, constructing a training set from scratch (\eg, CS-Bench and GTA) entails more in-depth research and extended experimental iterations for the system to discover effective strategies.

In Appendix~\ref{sec:appendix_addition_exp}, we provide the trajectories of performance scores across all experiments throughout the iterative process.

\begin{table*}[htbp]
\centering
\caption{Performance of the TREX framework for autonomous LLM fine-tuning on FT-Bench. We employ Qwen3-1.7B as the base model across all 10 tasks and compare Qwen3-Next-80B-Thinking and Gemini 3 Pro as the Researcher backend. To mitigate variance in task difficulty and metric scales, relative gains (in parentheses) are normalized to the performance gap between a strong reference (Qwen3-235B-Thinking) and the base model.}
\label{tab:exp_results_1}
\footnotesize
\renewcommand{\arraystretch}{1.2}
\begin{minipage}{\linewidth}
\centering
\begin{tabular}{L{0.1\textwidth} C{0.1\textwidth} C{0.12\textwidth} C{0.12\textwidth} C{0.22\textwidth} C{0.14\textwidth}}
\toprule
\multirow{2}{*}{\textbf{Task}} & \multirow{2}{*}{\textbf{Metric}$^{\dagger}$} & \textbf{Ref Model} & \textbf{Base Model} & \multicolumn{2}{c}{\textbf{TREX}} \\
\cmidrule(lr){5-6}
& & {\scriptbf{Qwen3-235B-2507}} & {\scriptbf{Qwen3-1.7B}} & {\scriptbf{w/ Qwen3-Next-80B-Thinking}} & {\scriptbf{w/ Gemini 3 Pro}}\\
\midrule
ACI-Bench    &Rouge-1& 0.240 & 0.205 & 0.260~\tinyit{(+157\%)} & 0.502~\tinyit{(+849\%)}\\
TOMG-Bench   &Val. $\&$ Acc. & 0.642 & 0.182 & 0.557~\tinyit{(+ 82\%)} & 0.681~\tinyit{(+108\%)}\\
oMeBench     &oMeScore& 0.283 & 0.198 & 0.392~\tinyit{(+228\%)} & 0.484~\tinyit{(+336\%)}\\
HoC          &Macro-F1& 0.645 & 0.462 & 0.896~\tinyit{(+237\%)} & 0.897~\tinyit{(+238\%)}\\
CS-Bench     &Acc.& 0.853 & 0.532 & 0.572~\tinyit{(+ 12\%)} & 0.581~\tinyit{(+ 15\%)}\\
OpenFinData  &Acc.& 0.833 & 0.494 & 0.688~\tinyit{(+ 57\%)} & 0.699~\tinyit{(+ 60\%)}\\
SST-2        &Acc.& 0.970 & 0.958 & 0.963~\tinyit{(+ 42\%)} & 0.972~\tinyit{(+117\%)}\\
EconlogicQA  &Acc.& 0.469 & 0.260 & 0.392~\tinyit{(+ 63\%)} & 0.454~\tinyit{(+ 93\%)}\\
GTA          &Acc.& 0.722 & 0.582 & 0.613~\tinyit{(+ 22\%)} & 0.652~\tinyit{(+ 50\%)}\\
LawBench     &Hybrid& 0.512 & 0.242 & 0.421~\tinyit{(+ 66\%)} & 0.409~\tinyit{(+ 62\%)}\\
\bottomrule
\end{tabular}
\par\raggedright
\scriptsize $^{\dagger}$ The numerical results of different metrics are normalized to facilitate cross-task comparability.
\end{minipage}
\end{table*}

\noindent\textbf{Comparison with Human finetune.}
Figure \ref{fig:score_trend_tomg} illustrates the performance growth curves of TREX across two benchmarks: TOMG-Bench and OpenFinData, providing a comparative analysis against improvements achieved by human experts via model fine-tuning.
On the TOMG-Bench task, we compare TREX against OpenMolIns-Large~\cite{li2024tomg}, the Llama3.1/3.2 variants fine-tuned by the TOMG-Bench authors on their specialized dataset. When fine-tuning the Qwen3-1.7B model, TREX delivers a substantial performance gain of 0.498, whereas OpenMolIns-Large achieves improvements of 0.189 and 0.139 on Llama3.1-8B and Llama3.2-1B, respectively.
Similar competitive advantages are observed in the OpenFinData task, where TREX yields a performance gain of 0.205 on Qwen3-1.7B. In comparison, the human expert solution FEVO~\cite{Pang2025FEVOFK} achieves a mere 0.025 improvement (FEVO-R32B-0) only applying RL on Qwen2.5-32B-Instruct, and reaches a 0.207 gain (FEVO-R32B) through a complex CPT-SFT-RL pipeline applied to Qwen2.5-32B base.
Collectively, these findings underscore TREX's competitiveness in enhancing large language model (LLM) performance through an end-to-end approach.

\begin{center}
    \centering
    \includegraphics[width=0.45\textwidth]{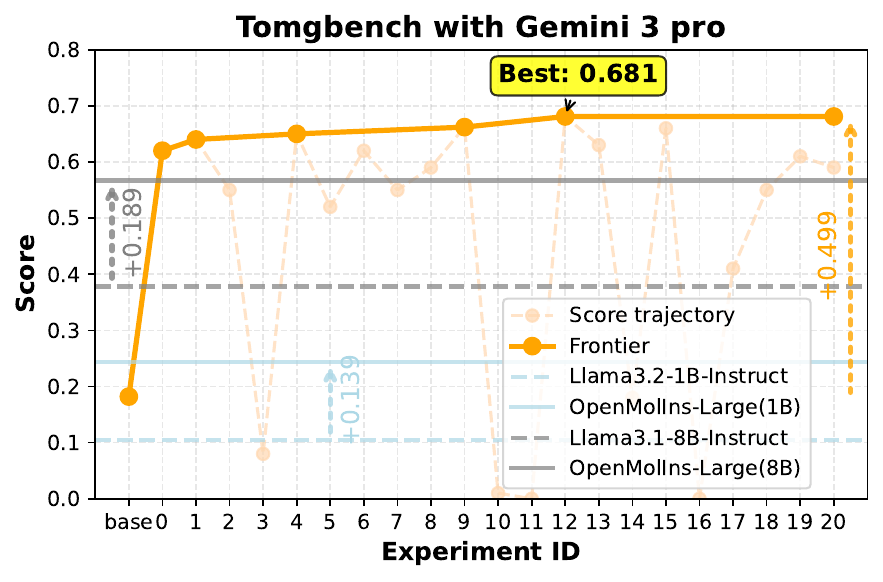}
     \includegraphics[width=0.45\textwidth]{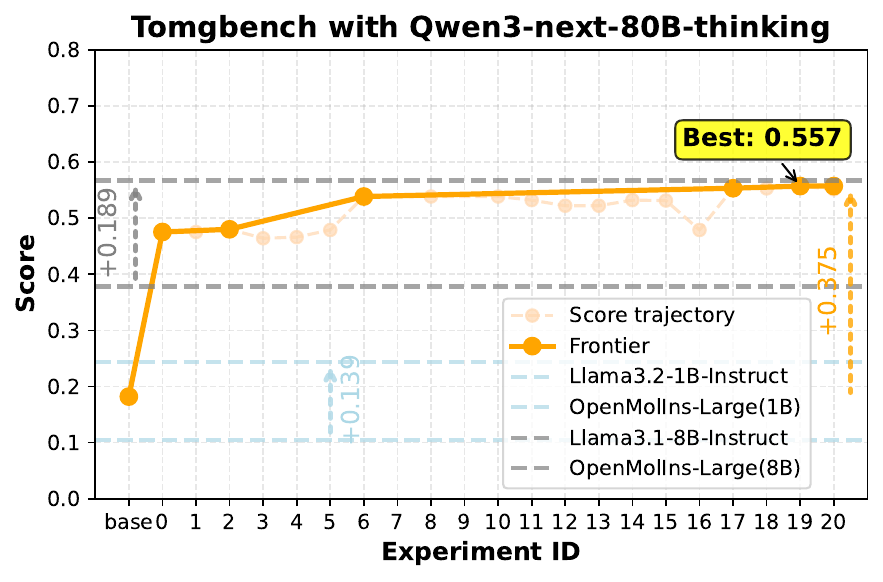} \\
     \includegraphics[width=0.45\textwidth]{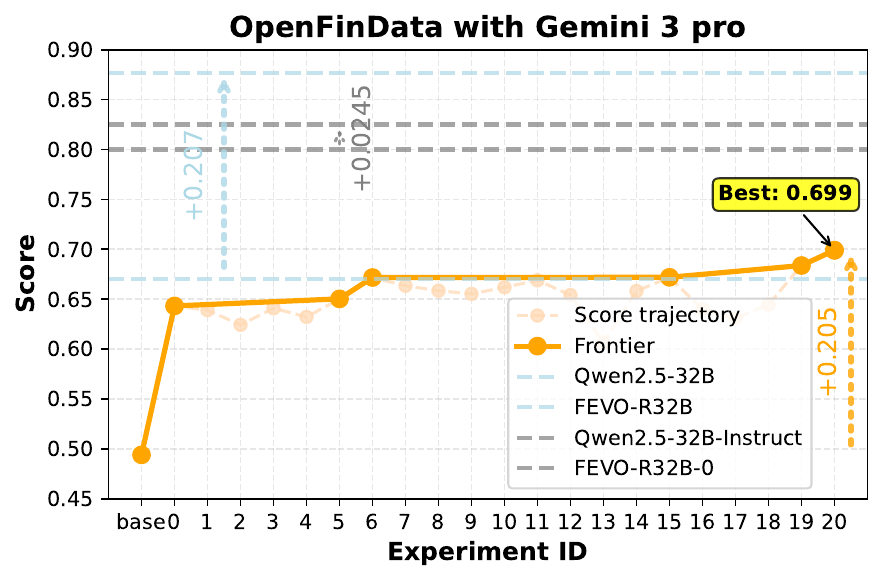}
     \includegraphics[width=0.45\textwidth]{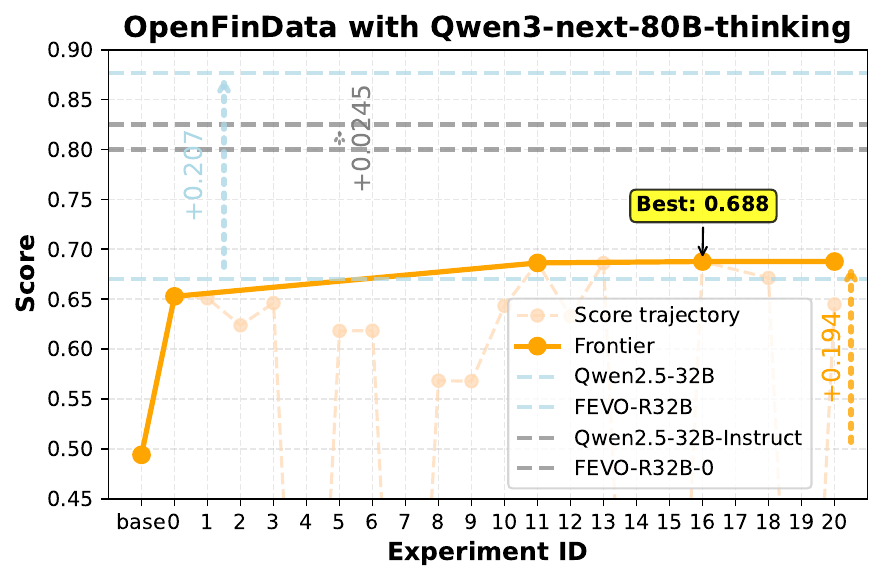}
    \captionof{figure}{Comparative performance trajectories of iterative experimental refinement using distinct LLM engines for the task  TOMG-Bench and OpenFinData. The left panel illustrates the score progression with Gemini 3 Pro as the Researcher, while the right panel depicts the corresponding progression with Qwen3-Next-80B-Thinking.}
     \label{fig:score_trend_tomg}
\end{center}

\noindent\textbf{Node Expansion Strategies}
As introduced in Section 3.2, the Researcher agent employs a coarse-to-fine approach, where high-level strategies are initially proposed to guide the generation of diverse experimental configurations. Table~\ref{tab:improvement_strategies} summarizes the frequency of different strategy types across all experimental trials. Evidently, TREX explores optimization strategies from various dimensions, encompassing both data processing and training methodologies. Furthermore, the Gemini 3 Pro backend outperforms the Qwen3-Next-80B-Thinking backend in terms of strategy diversity and execution success rate.

\begin{table}[!ht]
\caption{Frequency distribution of strategies proposed by TREX during the FT-Bench evaluation. (A: Attempts, S: Successes, I: Improvements)}
\label{tab:improvement_strategies}
\footnotesize
\centering
\renewcommand{\arraystretch}{1.2}
\begin{tabular}{L{0.3\textwidth} *{6}{C{0.05\textwidth}}}
\hline
\multirow{2}{*}{\textbf{Action Strategy}} & \multicolumn{3}{c}{\scriptbf{w/ Qwen3-Next-80B-Thinking}} & \multicolumn{3}{c}{\scriptbf{w/ Gemini 3 Pro}}\\
\cmidrule(lr){2-4} \cmidrule(lr){5-7}
& \textbf{A} & \textbf{S} & \textbf{I} & \textbf{A} & \textbf{S} & \textbf{I} \\
\hline
Establish Baseline        & 10  & 10  & -- & 10  & 10  & -- \\
Refine Data Pipeline      & 117 & 86  & 9  & 91  & 76  & 16 \\
Construct Synthetic Data  & 33  & 22  & 2  & 50  & 45  & 9  \\
Adjust Training Scheme    & 50  & 39  & 6  & 59  & 52  & 13 \\
\hline
\end{tabular}
\end{table}

%%%%%%%%%%%%%%%%%%%%%%%%%%%%%%%%%%%%%%%%%%%%%%%%%%%%%
\subsection{Ablation Studies}
%%%%%%%%%%%%%%%%%%%%%%%%%%%%%%%%%%%%%%%%%%%%%%%%%%%%%

In this section, we present comprehensive ablation studies to assess three pivotal design choices within the TREX framework: the tree-search strategies, the incorporation of AIDP, and the inclusion of bad-case analysis during experiment diagnostics. We evaluate these ablations on two representative tasks, OmeBench and GTA, using Gemini 3 Pro and the same default experimental settings as in Sec~\ref{sec:exp_setup}.

\noindent\textbf{Tree Search Strategies.}
We evaluate MCTS against two baseline strategies: (1) Greedy Best-First Search (GBFS), which expands the currently highest-scoring node; and (2) Sequential Expansion Search (SES), which iteratively expands from the previously selected node.
As illustrated in Figure~\ref{fig:search_strategies}, the MCTS strategy exhibits superior stability throughout the agent’s experimental trials, characterized by significantly lower fluctuations compared to baselines, coupled with a greater capacity for consistent performance gains over continuous experimental iterations.

\begin{center} 
\centering
\includegraphics[width=0.45\textwidth]{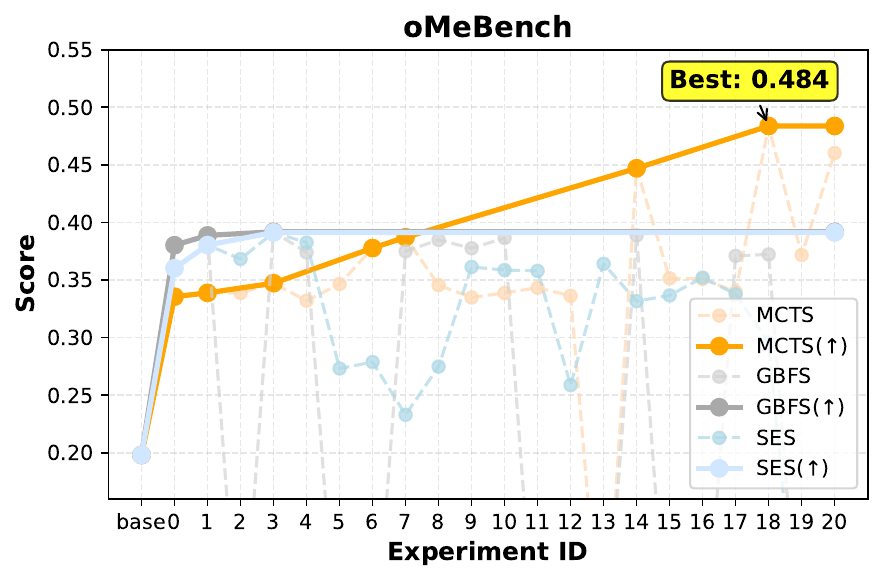} 
\includegraphics[width=0.45\textwidth]{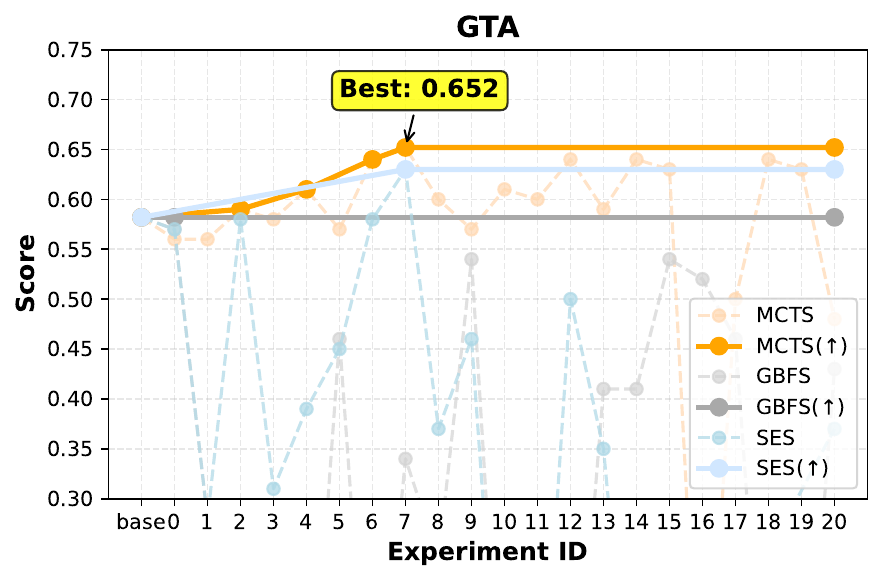}
\captionof{figure}{Performance of TREX across search strategies on oMeBench and GTA tasks. The dashed line represents the score trajectory, while the solid line with $\uparrow$ represents the frontier of the score trajectory.}
\label{fig:search_strategies}
\end{center}

\noindent\textbf{Incorporation of AIDP.} 
We further compare the system’s ability to construct training data pipelines with and without the AIDP library. The experimental results in Figure~\ref{fig:aidp_strategies} demonstrate that, in the absence of AIDP tools, the resulting performance improvements are substantially lower than those achieved when AIDP tools are employed. Moreover, an examination of the score trajectories reveals that experiments conducted without AIDP support are more susceptible to interruptions in subsequent training, primarily due to data-processing failures during the experimental procedure.

\begin{center}
\centering
\includegraphics[width=0.45\textwidth]{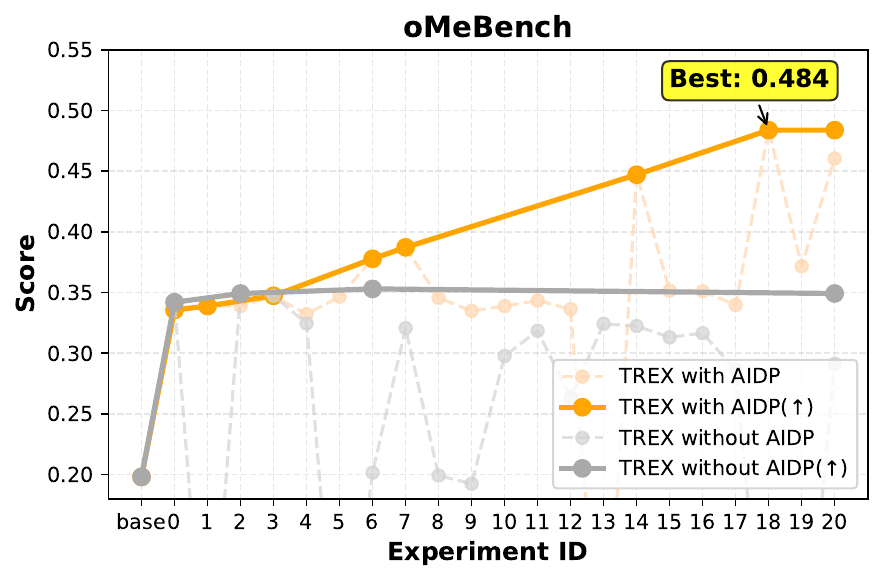}
\includegraphics[width=0.45\textwidth]{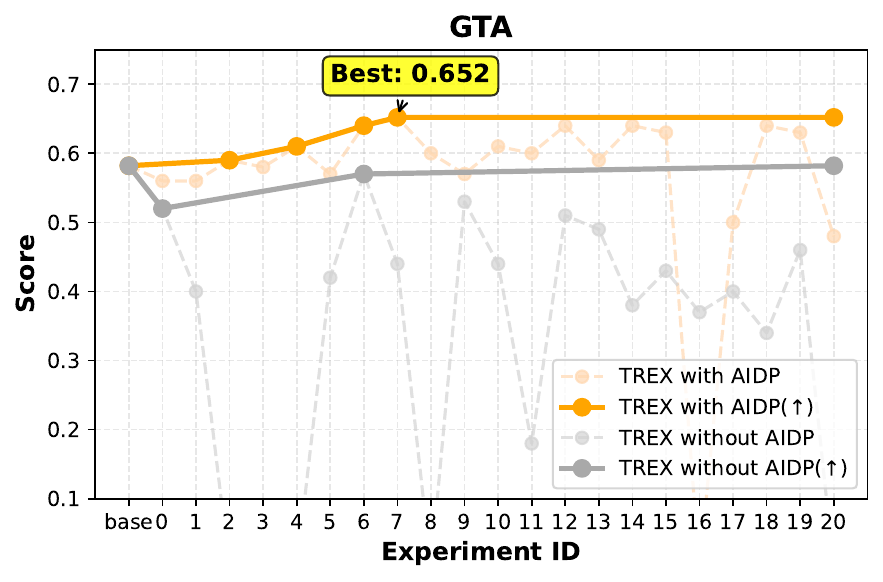}
\captionof{figure}{Performance of TREX  with versus without AIDP tools on oMeBench and GTA tasks. The dashed line represents the score trajectory, while the solid line with $\uparrow$ represents the frontier of the score trajectory.}
\label{fig:aidp_strategies}
\end{center}

\noindent\textbf{Bad-case Analysis.}
Additionally, we conduct controlled ablation experiments to evaluate the effectiveness of incorporating bad-case analysis into the experimental diagnostics process, as illustrated in~\ref{fig:badcase_strategies}. The results indicate that observing and analyzing bad cases enables the system to iterate on experiments more effectively, ultimately leading to improved final performance scores.

\begin{center}
\centering
\includegraphics[width=0.45\textwidth]{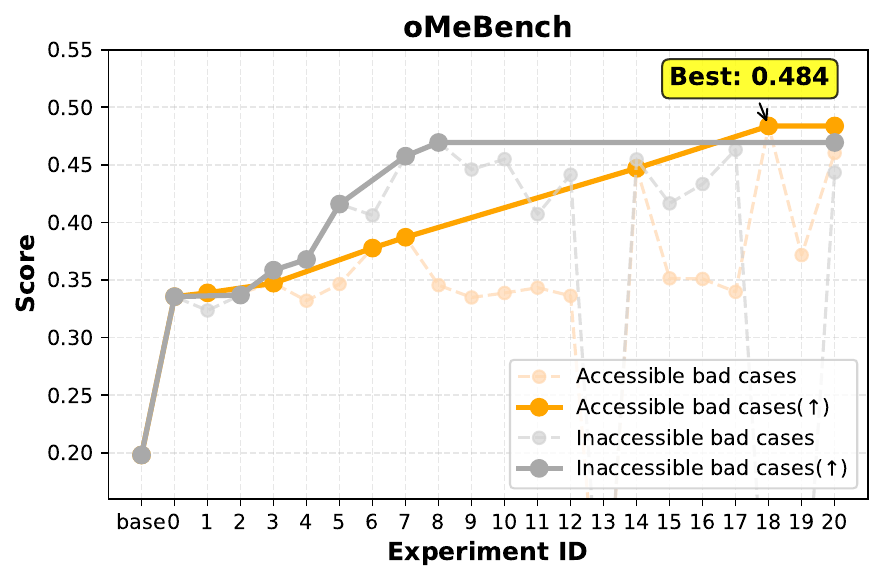}
\includegraphics[width=0.45\textwidth]{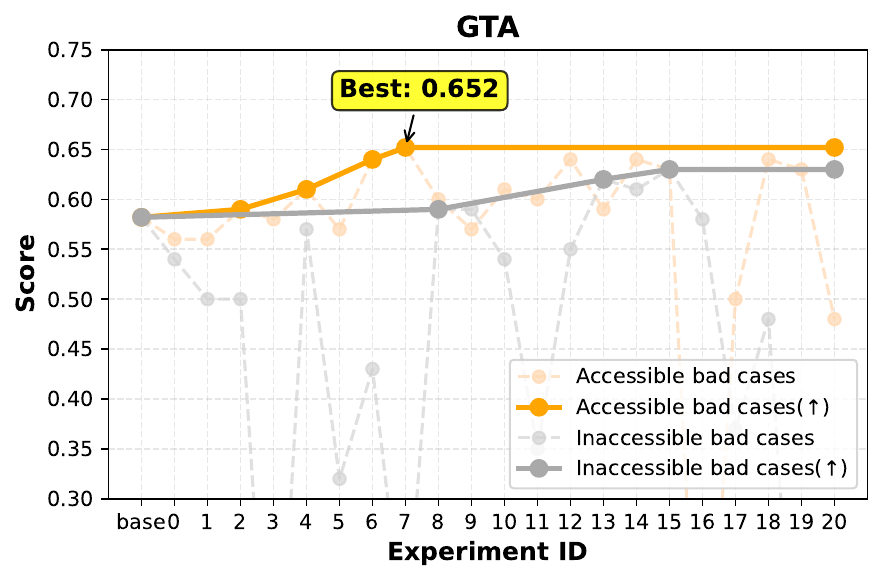}
\captionof{figure}{Performance of TREX  with accessible versus inaccessible bad cases on oMeBench and GTA tasks. The dashed line represents the score trajectory, while the solid line with $\uparrow$ represents the frontier of the score trajectory.}
\label{fig:badcase_strategies}
\end{center}

%% file: sections/6_conclusion.tex
\section{Conclustion}
\label{sec:conclusion}

We present TREX, a multi-agent framework that automates the full lifecycle of LLM fine-tuning via coordinated planning, execution, and iterative refinement. By formulating training optimization as a tree-based search problem and leveraging MCTS, TREX efficiently explores the open-ended space of training strategies under constrained computational budgets. We further introduce FT-Bench, a dedicated benchmark for evaluating agent-driven LLM training systems. Empirical results show that TREX consistently improves model performance across diverse tasks, in several cases matching or surpassing expert-designed pipelines. These findings highlight the potential of agent-based systems as a scalable paradigm for automated AI research.

%% file: sections/appendix.tex
\section{Appendix}

%%%%%%%%%%%%%%%%%%%%%%%%%%%%%%%%%%%%%
\subsection{Details of the tasks in FT-Bench}
\label{sec:appendix_benchmark_detail}
%%%%%%%%%%%%%%%%%%%%%%%%%%%%%%%%%%%%%

\begin{itemize} 
    \item \textbf{ACI-Bench}~\cite{yim2023aci}: This task assesses the capability of large language models to generate AI-assisted medical notes based on clinical visit dialogues. The primary evaluation metric is Rouge-1. The dataset is partitioned into 335 training samples, 600 test samples, and 100 validation samples.
        
    \item \textbf{TOMG-Bench}~\cite{li2024tomg}: This task evaluates the performance of large language models in text-based open molecule generation. The evaluation metric is a weighted score combining Validity and Accuracy , formulated as \(\text{Score} = 0.4 \times \text{Validity} + 0.6 \times \text{Accuracy}\). The \textit{large} split, comprising \(90{,}000\) samples, is used as training data. For each test task, \(300\) samples are selected, yielding a total of \(2{,}700\) samples, which are randomly divided into \(500\) validation samples and \(2{,}200\) test samples.

  \item \textbf{oMeBench}~\cite{xu2025omebench}: This task evaluates the reasoning capabilities of large language models in organic chemical reaction mechanism analysis. The evaluation metric is a weighted score based on oMeScore, which combines SMILES Validity (\(V\)), Logical Fidelity Score (\(L\)), oMeS-total (\(S_{\text{tot}}\)), and oMeS-partial (\(S_{\text{part}}\)), and is formulated as
\[
\text{Score} = 0.2 \times V + 0.2 \times L + 0.3 \times S_{\text{tot}} + 0.3 \times S_{\text{part}}.
\]
From the original dataset, 293 samples from the training split are held out as the validation set, while the remaining 2,200 samples are used for training.

    \item \textbf{Hoc}~\cite{baker2016automatic}: This task assesses the capability of large language models to classify cancer-related scientific literature. The primary evaluation metric is Macro-F1. The dataset is partitioned into 1,108 training samples, 315 test samples, and 157 validation samples.
    
    \item \textbf{CS-Bench}~\cite{song2024cs}: This task evaluates the performance of large language models on bilingual (Chinese-English) computer science questions. The evaluation metric is accuracy. Only multiple-choice and true/false questions from the original dataset are retained. The resulting split contains 1,711 test samples and 194 validation samples.
    
    \item \textbf{OpenFinData}~\cite{eastmoney2023openfindata}: This task assesses the capability of large language models in authentic, comprehensive, and domain-specialized financial scenarios. The evaluation metric is the average accuracy across all sub-tasks. Following the OpenCompass configuration, 650 test samples are selected from 8 sub-tasks.
    
    \item \textbf{SST-2}~\cite{socher2013recursive}: This task evaluates the performance of large language models in movie review sentiment classification. The evaluation metric is accuracy. The dataset is divided into 67,349 training samples, 1,821 test samples, and 872 validation samples.
    
    \item \textbf{EconlogicQA}~\cite{quan2024econlogicqa}: This task assesses the logical reasoning ability of large language models in complex scenarios involving economics, business, and supply chain management. The evaluation metric is accuracy. The dataset is partitioned into 260 training samples, 130 test samples, and 130 validation samples.
    
    \item \textbf{GTA}~\cite{wang2024gta}: This task evaluates the tool-use capabilities of LLM-based agents in real-world scenarios. The evaluation metric is accuracy. The benchmark consists of 229 test samples.
    
    \item \textbf{LawBench}~\cite{fei2024lawbench}: This task assesses the legal cognitive capabilities of large language models across 20 tasks covering legal knowledge memorization, comprehension, and application. The evaluation metric is an average score aggregated from task-specific metrics such as Accuracy, ROUGE-L, F1, etc. For each sub-task, 250 instances are randomly sampled, resulting in a total of 5,000 test samples.
\end{itemize}

%%%%%%%%%%%%%%%%%%%%%%%%%%%%%%%%%%%%%
\subsection{AIDP Package}
\label{sec:appendix_aidp}
%%%%%%%%%%%%%%%%%%%%%%%%%%%%%%%%%%%%%

Equipping LLM agents with data-processing tools is essential for experimental robustness, reproducibility, and efficient large-scale data manipulation. In this work, we propose AIDP, a suite of high-performance operators built on the HuggingFace Datasets ecosystem. By exposing the Python signatures of these tools to the agent, TREX can autonomously orchestrate and implement complex data pipelines via function calls within generated scripts. Table~\ref{tab:aidp} details the available tools.

\begin{table*}[ht]
\centering
\small % 适当缩小字号以适应版面
\renewcommand{\arraystretch}{1.2} % 增加行高，提升阅读体验
\scalebox{0.9}{
\begin{tabular}{l l p{8cm}} % p{8cm} 自动换行
\toprule
\textbf{Category} & \textbf{Operator} & \textbf{Function Description} \\
\midrule
\multirow{2}{*}{\textit{Loader}} 
& \texttt{load\_local\_dataset} & Imports datasets from local storage. \\
& \texttt{load\_remote\_dataset} & Retrieves datasets from remote repositories (e.g., Hugging Face). \\
\midrule
\multirow{2}{*}{\textit{Scorer}} 
& \texttt{compute\_perplexity} & Calculates text perplexity. \\
& \texttt{score\_dataset\_with\_llm} & Employs an LLM as a judge to evaluate data. \\
\midrule
\multirow{3}{*}{\textit{Generator}} 
& \texttt{generate\_text\_embeddings} & Computes dense vector representations for dataset entries. \\
& \texttt{generate\_dataset\_with\_llm} & Synthesizes Instruction-Response pairs from seed data. \\
& \texttt{generate\_preference\_dataset} & Constructs preference pairs based on custom ranking logic. \\
\midrule
\multirow{4}{*}{\textit{Filter}} 
& \texttt{deduplicate\_by\_text\_hash} & Eliminates redundant samples via exact hash matching. \\
& \texttt{select\_by\_filter} & Filters data instances based on a user-defined boolean predicate. \\
& \texttt{select\_by\_score} & Selects top-k samples according to a specific scoring metric. \\
& \texttt{select\_by\_random} & Performs random sampling from the dataset. \\
\bottomrule
\end{tabular}
}
\vspace{0.1cm}
\caption{\textbf{Overview of Primitive Operators in AIDP.} The toolkit provides a comprehensive set of operators for data ingestion, formation, analysis, synthesis, and selection.}
\label{tab:aidp}
\end{table*}

%%%%%%%%%%%%%%%%%%%%%%%%%%%%%%%%%%%%%
\subsection{Additional Experiment Results}
\label{sec:appendix_addition_exp}
%%%%%%%%%%%%%%%%%%%%%%%%%%%%%%%%%%%%%

\begin{figure*}[htbp]
    \centering
    % 相对宽度（推荐）
    \includegraphics[width=0.45\textwidth]{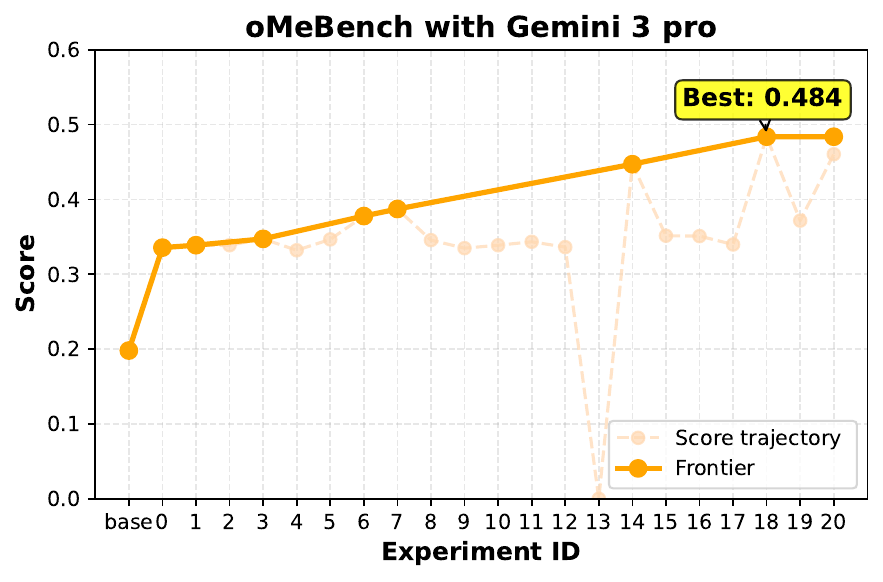}
     \includegraphics[width=0.45\textwidth]{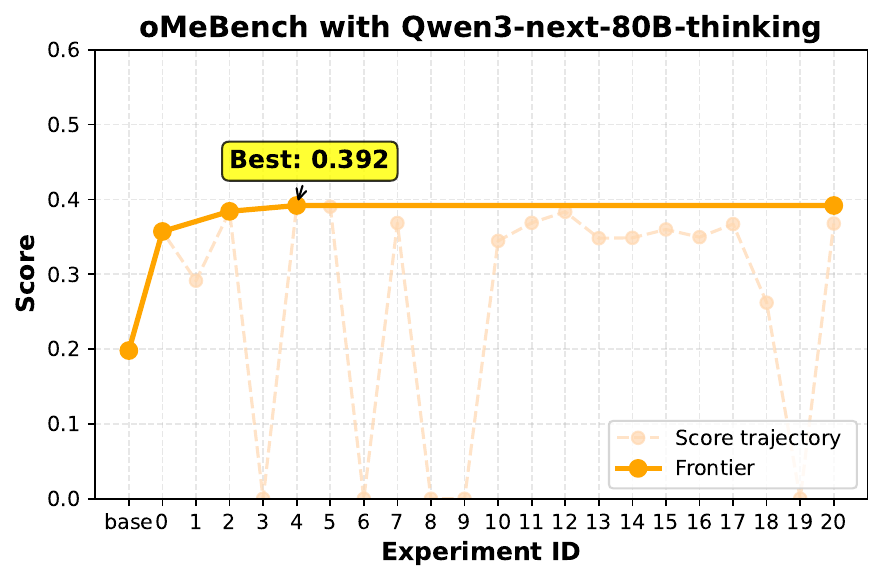} \\
      \includegraphics[width=0.45\textwidth]{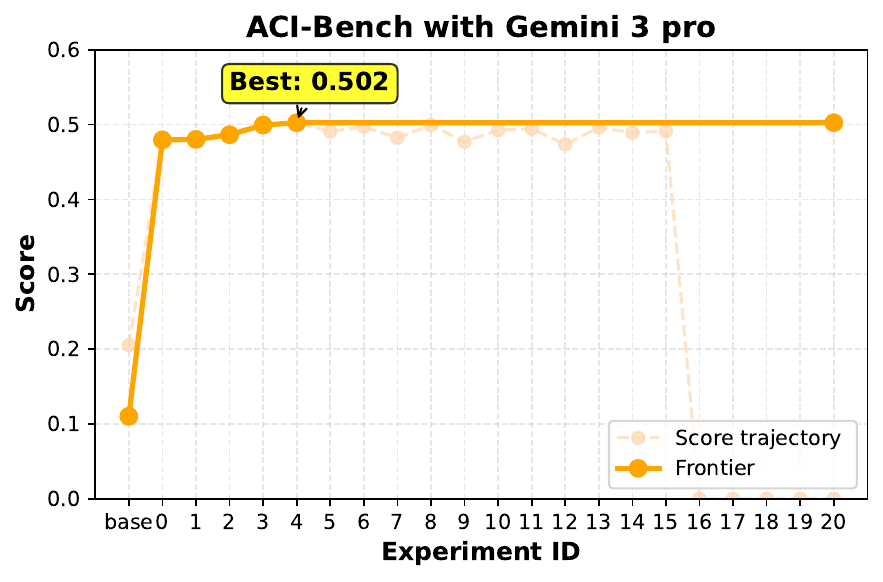}
     \includegraphics[width=0.45\textwidth]{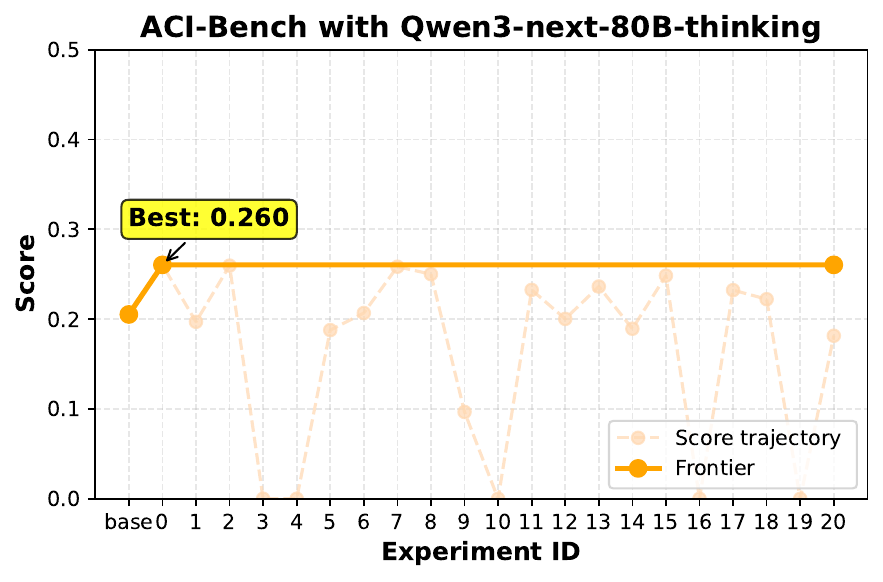}\\
      \includegraphics[width=0.45\textwidth]{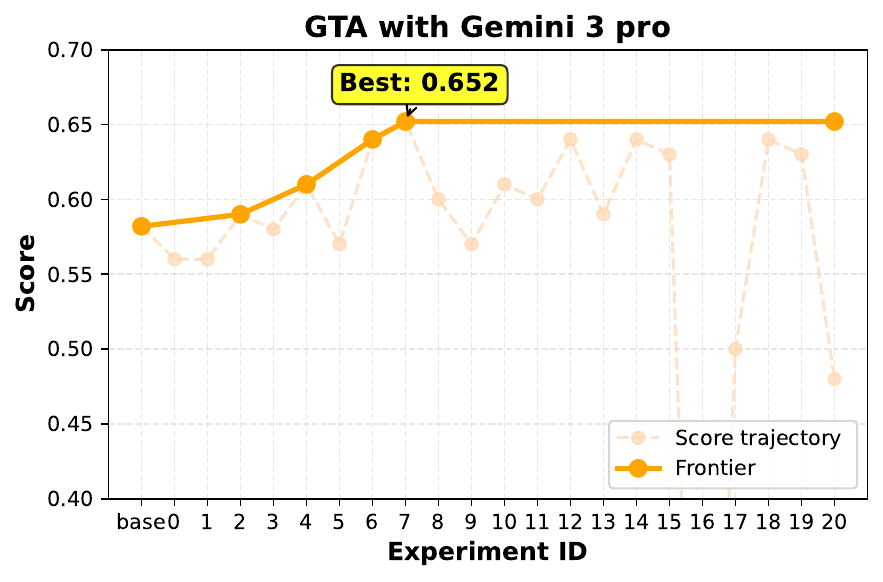}
     \includegraphics[width=0.45\textwidth] {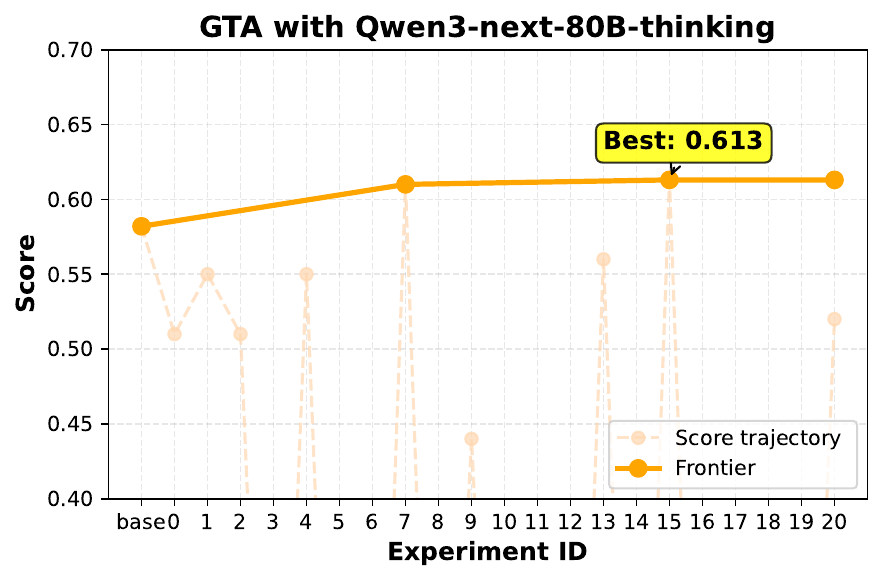} \\
      \includegraphics[width=0.45\textwidth]{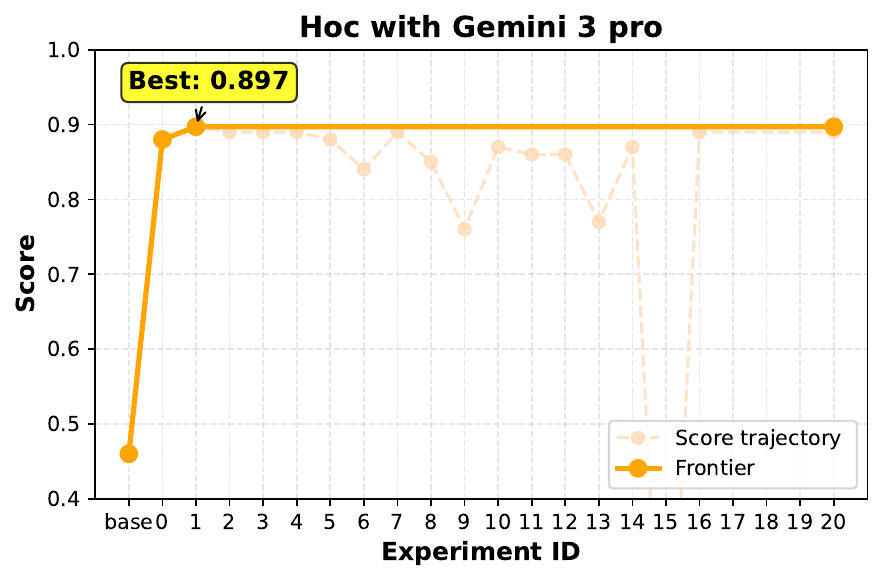}
     \includegraphics[width=0.45\textwidth]{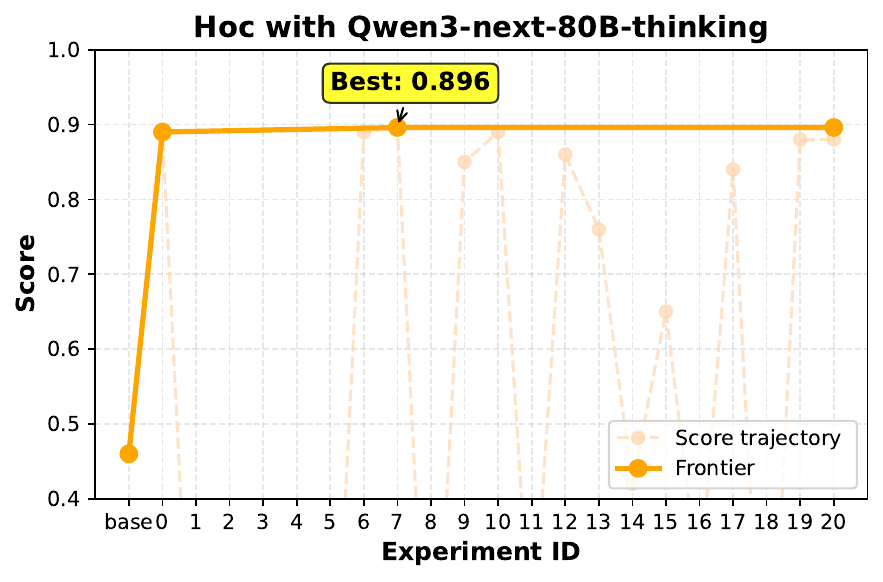} \\
    % 或绝对宽度
    % \includegraphics[width=10cm]{experiment_score_trend.pdf}
    \caption{Comparative performance trajectories of iterative experimental refinement using distinct LLM engines for tasks oMeBench, ACI-Bench, GTA, and HOC. The left panel illustrates the score progression with Gemini 3 pro as the Researcher, while the right panel depicts the corresponding progression with Qwen3-next-80b-thinking.}
    \label{fig:addition_exp_trajectories_p1}
\end{figure*}

\begin{figure*}[htbp]
    \centering
  
    % 相对宽度（推荐）
   
      \includegraphics[width=0.45\textwidth]{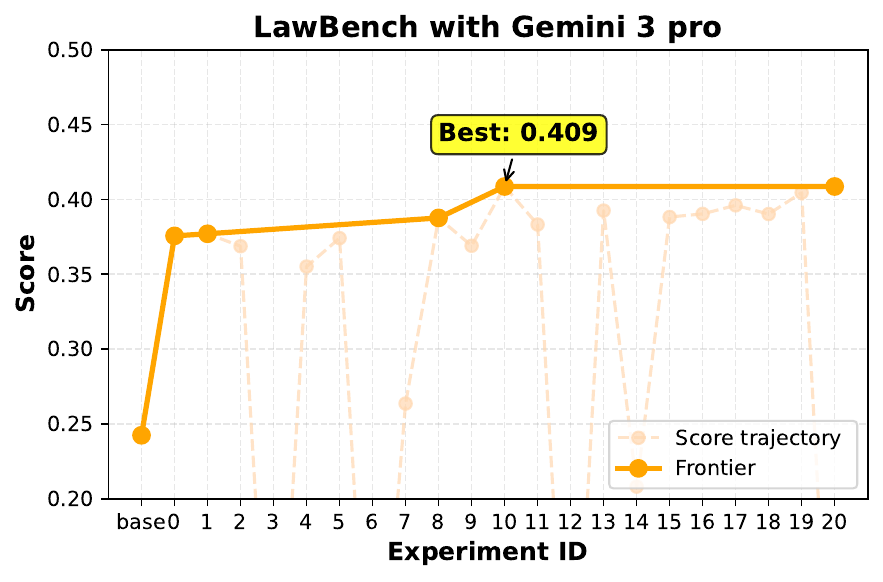}
     \includegraphics[width=0.45\textwidth]{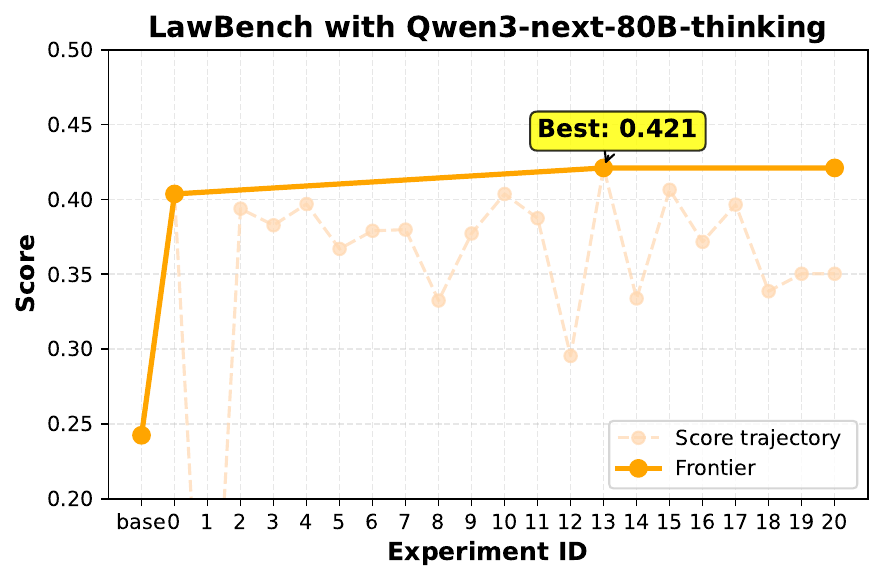} \\
      \includegraphics[width=0.45\textwidth]{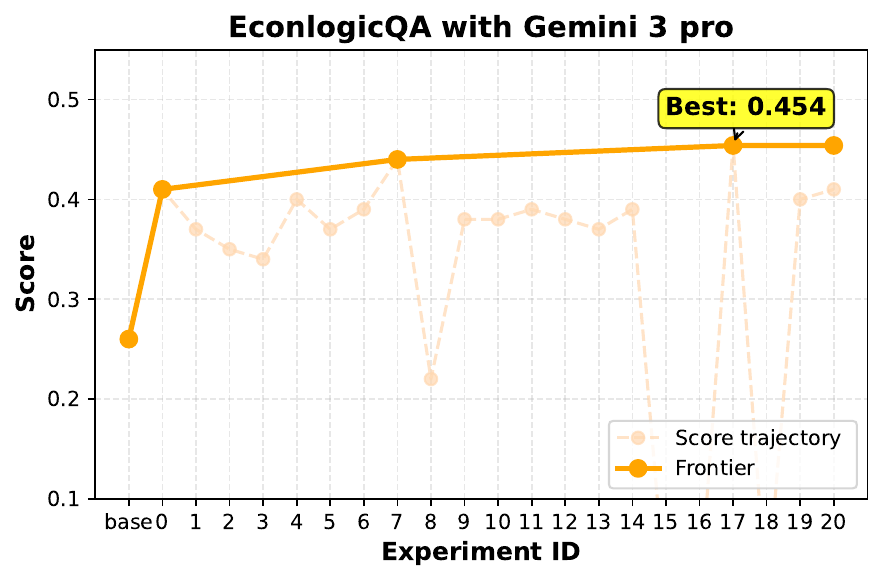}
     \includegraphics[width=0.45\textwidth]{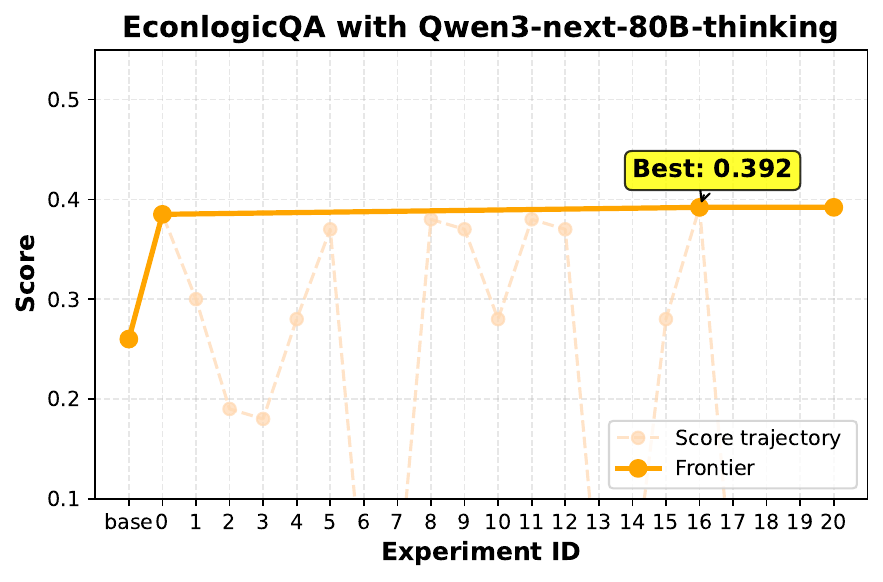}\\
      \includegraphics[width=0.45\textwidth]{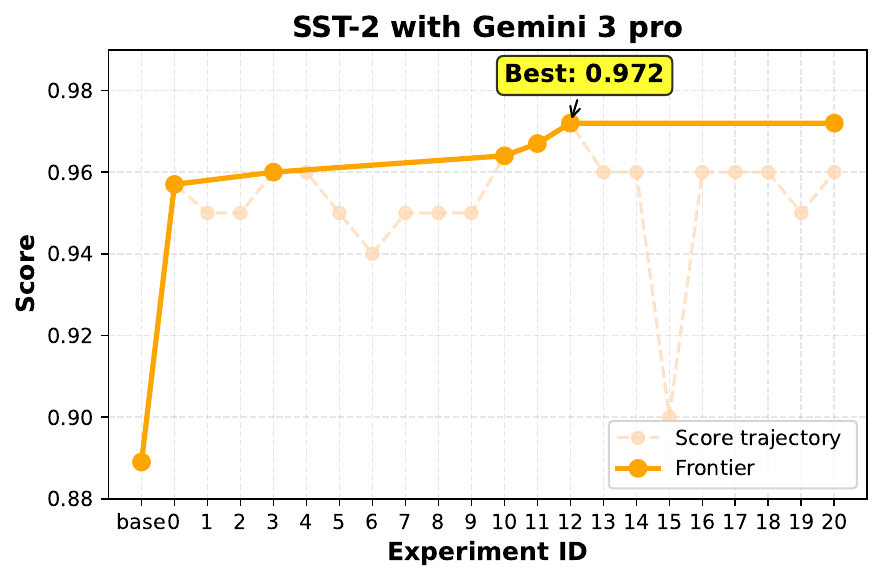}
     \includegraphics[width=0.45\textwidth]{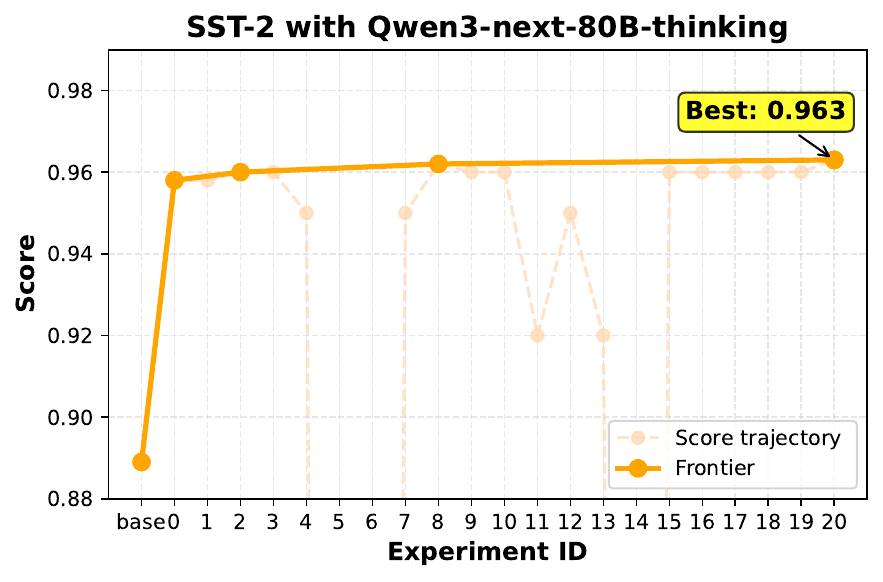} \\
     \includegraphics[width=0.45\textwidth]{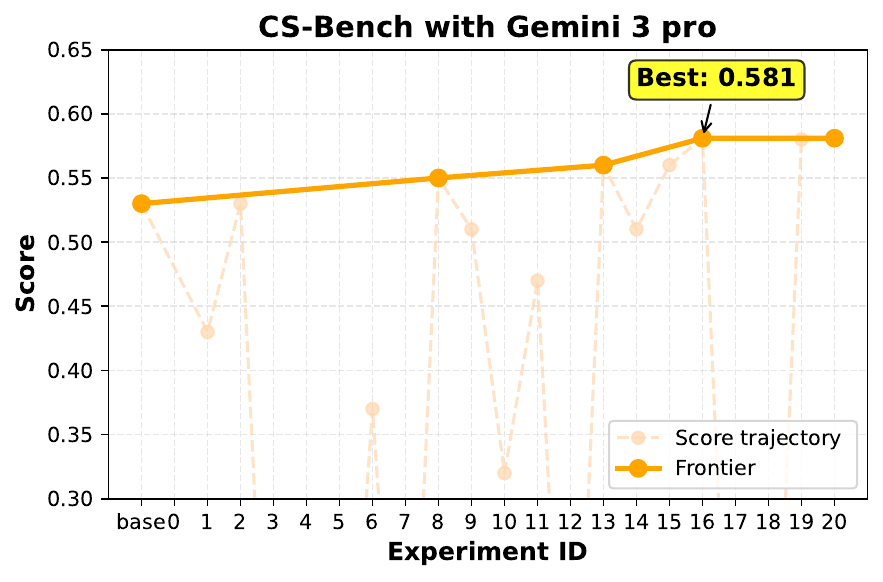}
     \includegraphics[width=0.45\textwidth]{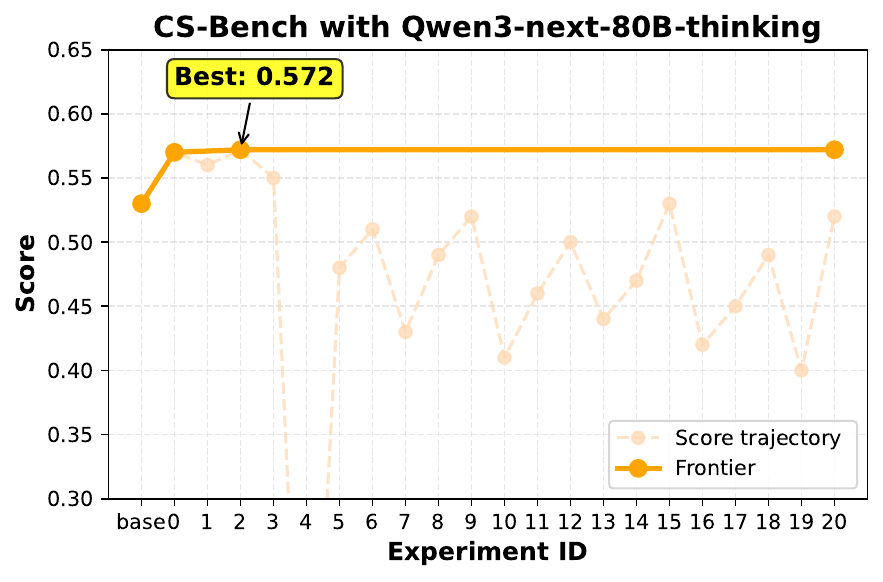}
    % 或绝对宽度
    % \includegraphics[width=10cm]{experiment_score_trend.pdf}
      \vspace{-0.2cm}
    \caption{Comparative performance trajectories of iterative experimental refinement using distinct LLM engines for the task LawBench and EconlogicQA, SST-2 and CS-Bench. The left panel illustrates the score progression with Gemini 3 pro as the Researcher, while the right panel depicts the corresponding progression with Qwen3-next-80b-thinking.}
    \label{fig:addition_exp_trajectories_p2}
\end{figure*}

Figure~\ref{fig:addition_exp_trajectories_p1} and Figure~\ref{fig:addition_exp_trajectories_p2} illustrate the performance trajectories of TREX across all tasks in FT-Bench.

%%%%%%%%%%%%%%%%%%%%%%%%%%%%%%%%%%%%%
\subsection{Experiment Plan Examples}
\label{sec:appendix_exp_plans}
%%%%%%%%%%%%%%%%%%%%%%%%%%%%%%%%%%%%%
 
We showcase the experimental plans for three rounds of the experiments (Experiment-1, Experiment-4, and Experiment-12) in the molecule generation task using the TOMG-Bench. The overall experimental trajectory is shown in Fig. \ref{fig:teaser}.

\begin{mybox}{Experiment-1 Plan: Establishing a Baseline}
    \textbf{[Experiment Mission]}
    
    Determine the training hyperparameters (data volume, batch size, learning rate, number of training steps, and LoRA rank) via grid search to establish a baseline model for the chemical molecule generation task, validate the feasibility of the experimental pipeline, and identify a reasonable range of parameters.

    \textbf{[Training Data]} 
    
    Utilizing the user-provided TOMG-Bench training dataset (path: $/shared\_data/user\_data/tomgbench/train.jsonl$), which has been formatted in accordance with the LLaMA Factory standard and contains 90,000 instruction fine-tuning samples, the data processing procedure is as follows:

    \begin{itemize}[noitemsep, topsep=0pt]
        \item A specified number of samples (10K/25K/50K) are randomly extracted from the original dataset to generate new JSONL files.
        \item The saved path is: \texttt{<workspace>/datasets/train\_\{n\}k.jsonl} (where $n$ represents the number of samples in thousands).
        \item Selection rationale: This dataset is specifically designed for the TOMG-Bench task, containing instruction-response pairs for three types of tasks—molecule editing, optimization, and generation. Furthermore, it does not overlap with the evaluation set (TOMG-Bench test set), as the training set is sourced from OpenMolIns, while the test set is independently sampled.
        \item Samples: 10K, 25K, 50K (each training run does not exceed 50,000 samples).
    \end{itemize}

    \textbf{[Core Training Strategy]}
    \begin{itemize}
    \item Base model: \texttt{Qwen/Qwen3-1.7B}
    \item finetuning\_type: lora
    \item cutoff\_len: 2048
    \item num\_gpus: 4
    \item Other parameters are adjusted according to experimental setups, as below:
    \begin{itemize}
        \item \textbf{Data Scale Experiments (3 groups)}:
        \begin{itemize}
            \item 10K samples: per\_device\_train\_batch\_size = 2, gradient\_accumulation\_steps = 8, learning\_rate = 3e-4, num\_train\_epochs = 2, lora\_rank = 8
            \item 25K samples: same as above
            \item 50K samples: same as above
        \end{itemize}
        
        \item \textbf{Batch Size Experiments (2 groups)}:
        \begin{itemize}
            \item per\_device\_train\_batch\_size = 1, gradient\_accumulation\_steps = 16 (maintaining total batch size = 64)
            \item per\_device\_train\_batch\_size = 4, gradient\_accumulation\_steps = 4 (maintaining total batch size = 64)
            \item Other parameters: data scale = 50K, learning\_rate = 3e-4, num\_train\_epochs = 2, lora\_rank = 8
        \end{itemize}
        
        \item \textbf{Learning Rate Experiments (2 groups)}:
        \begin{itemize}
            \item learning\_rate = 1e-4
            \item learning\_rate = 5e-4
            \item Other parameters: data scale = 50K, per\_device\_train\_batch\_size = 2, gradient\_accumulation\_steps = 8, num\_train\_epochs = 2, lora\_rank = 8
        \end{itemize}
        
        \item \textbf{Training Epochs Experiments (2 groups)}:
        \begin{itemize}
            \item num\_train\_epochs = 1
            \item num\_train\_epochs = 3
            \item Other parameters: data scale = 50K, per\_device\_train\_batch\_size = 2, gradient\_accumulation\_steps = 8, learning\_rate = 3e-4, lora\_rank = 8
        \end{itemize}
        
        \item \textbf{LoRA Rank Experiments (2 groups)}:
        \begin{itemize}
            \item lora\_rank = 16
            \item lora\_rank = 32
            \item Other parameters: data scale = 50K, per\_device\_train\_batch\_size = 2, gradient\_accumulation\_steps = 8, learning\_rate = 3e-4, num\_train\_epochs = 2
        \end{itemize}
    \end{itemize}
    \item Total number of experiments: $3 + 2 + 2 + 2 + 2 = 11$ groups.
\end{itemize}
\end{mybox}

\begin{mybox}{Experiment-4 Plan: Improving Generalization via Uniform QED Distribution}
    \textbf{[Experiment Mission]}\\
    To address the reduced generalization ability observed in the third-round experiment due to the narrow QED distribution of the training data, this experiment restructures the training set to contain a uniform distribution of low-, medium-, and high-QED samples. The goal is to improve the model's adaptability to diverse QED constraints by covering three QED intervals: 0.0--0.3, 0.3--0.7, and 0.7--1.0. This design ensures that the model can handle generation and optimization tasks across different QED ranges, while mitigating overfitting and degraded task adaptability caused by training on a single interval.

    \textbf{[Training Data]}\\
    Utilizing the HuggingFace dataset \texttt{edmanft/zinc250k} (no split or configuration specification is required, as the dataset has no sub-configurations), the data processing procedure is as follows:
    \begin{itemize}[noitemsep, topsep=0pt]
        \item The dataset is loaded using the \texttt{datasets} library.
        \item Samples are divided into three groups according to their QED values:
        \begin{itemize}[noitemsep, topsep=0pt]
            \item Low QED: $0.0 \leq \mathrm{QED} < 0.3$
            \item Medium QED: $0.3 \leq \mathrm{QED} < 0.7$
            \item High QED: $0.7 \leq \mathrm{QED} \leq 1.0$
        \end{itemize}
        \item A total of 10,000 samples are constructed by randomly sampling 3,333 instances from each interval.
        \item Each sample is transformed into a JSONL record in the instruction fine-tuning format, with the following structure:
        \begin{itemize}[noitemsep, topsep=0pt]
            \item \textbf{system content}: \texttt{"You are working as an assistant of a chemist user. Please follow the instruction of the chemist and generate a molecule that satisfies the requirements of the chemist user. You could think step by step, but your final response must contain the answer with SMILES string."}
            \item \textbf{user content}: a QED-conditioned instruction corresponding to the interval:
            \begin{itemize}[noitemsep, topsep=0pt]
                \item Low QED: \texttt{"Generate a molecule with QED value between 0.0 and 0.3"}
                \item Medium QED: \texttt{"Generate a molecule with QED value between 0.3 and 0.7"}
                \item High QED: \texttt{"Generate a molecule with QED value between 0.7 and 1.0"}
            \end{itemize}
            \item \textbf{assistant content}: the \texttt{smiles} field of the corresponding sample
        \end{itemize}
        \item Save the processed file to: \texttt{<workspace>/datasets/train\_qed\_uniform\_10k.jsonl}
        \item Selection rationale: The \texttt{zinc250k} dataset provides explicit QED annotations, making it well suited for constructing QED-related molecular generation tasks. The uniformly distributed sampling strategy covers the full QED range and helps avoid the generalization issues caused by a single-interval training distribution. In addition, the dataset size is moderate (10,000 samples), satisfying the training data constraint ($\leq 50{,}000$ samples). There is also no overlap with the TOMG-Bench test set, which contains only 9,000 independently sampled molecules from ZINC-250K.
    \end{itemize}

    \textbf{[Core Training Strategy]}
    \begin{itemize}
        \item Base model: \texttt{/<workspace>/0122205851-exp\_1/models/last} (the best-performing full fine-tuning model from the second-round experiment)
        \item finetuning\_type: full
        \item learning\_rate: \texttt{2e-5}
        \item num\_gpus: 4
        \item per\_device\_train\_batch\_size: 1
        \item gradient\_accumulation\_steps: 16
        \item num\_train\_epochs: 2
        \item cutoff\_len: 2048
        \item warmup\_ratio: 0.03
        \item max\_grad\_norm: 1.0
        \item work\_dir: \texttt{<workspace>}
        \item dataset: \texttt{<workspace>/datasets/train\_qed\_uniform\_10k.jsonl}
    \end{itemize}
\end{mybox}

\begin{mybox}{Experiment-12 Plan: Improving Instruction Naturalness via Real Chemical Descriptions}
    \textbf{[Experiment Mission]}
    
    To address the degradation in model generalization observed in Experiment 6 due to manually constructed, overly rigid instruction templates, this experiment introduces molecule descriptions from real chemical literature via the ChEBI-20-MM dataset as natural-language instructions, thereby reducing the mechanical nature of synthetic prompts. In addition, different mixing ratios between ChEBI-20-MM and TOMG-Bench training data are explored to balance the benefits of realistic, open-ended chemical descriptions and task-specific instructions, with the goal of improving the model's ability to understand diverse chemical constraints.

    \textbf{[Training Data]}
    
    Two training data sources are used in this experiment:
    \begin{itemize}[noitemsep, topsep=0pt]
        \item \textbf{ChEBI-20-MM dataset}: The dataset is obtained from HuggingFace under \texttt{liupf/ChEBI-20-MM}, with the split explicitly set to \texttt{train} (the dataset provides \texttt{train}/\texttt{validation}/\texttt{test} splits, so the training split must be clearly specified). This split contains 26,402 samples. Each sample includes a \texttt{description} field (a natural-language molecular description, e.g., ``epoxy(hydroxy)icosatrienoate'') and a \texttt{SMILES} field.
        
        The data are converted into JSONL format following the LLaMA Factory convention:
        \begin{itemize}[noitemsep, topsep=0pt]
            \item system role: \texttt{"You are working as an assistant of a chemist user. Please follow the instruction of the chemist and generate a molecule that satisfies the requirements of the chemist user. You could think step by step, but your final response must contain the answer with SMILES string."}
            \item user role: directly uses the content of the \texttt{description} field
            \item assistant role: directly uses the content of the \texttt{SMILES} field
        \end{itemize}
        The processed file is saved to: \texttt{<workspace>/datasets/chebi\_train.jsonl}
        
        \item \textbf{TOMG-Bench training data}: The preprocessed file is directly reused from \texttt{/shared\_data/user\_data/tomgbench/train.jsonl} and copied to \texttt{<workspace>/datasets/tomgbench\_train.jsonl}. This file contains 90,000 formatted instruction-response pairs.
        
        \item \textbf{Data mixing schemes} (with the total number of samples controlled to be no more than 50,000 in each run):
        \begin{itemize}[noitemsep, topsep=0pt]
            \item \textbf{Scheme 1 (100\% ChEBI)}: use only \texttt{chebi\_train.jsonl} (26,402 samples); saved as \texttt{<workspace>/datasets/chebi\_100.jsonl}
            \item \textbf{Scheme 2 (70\% ChEBI + 30\% TOMG)}: use all 26,402 samples from \texttt{chebi\_train.jsonl} and randomly sample 23,598 samples from \texttt{tomgbench\_train.jsonl}, then merge them into 50,000 samples; saved as \texttt{<workspace>/datasets/chebi70\_tomg30.jsonl}
            \item \textbf{Scheme 3 (40\% ChEBI + 60\% TOMG)}: randomly sample 20,000 samples from \texttt{chebi\_train.jsonl} and 30,000 samples from \texttt{tomgbench\_train.jsonl}, then merge them into 50,000 samples; saved as \texttt{<workspace>/datasets/chebi40\_tomg60.jsonl}
            \item \textbf{Scheme 4 (20\% ChEBI + 80\% TOMG)}: randomly sample 10,000 samples from \texttt{chebi\_train.jsonl} and 40,000 samples from \texttt{tomgbench\_train.jsonl}, then merge them into 50,000 samples; saved as \texttt{<workspace>/datasets/chebi20\_tomg80.jsonl}
        \end{itemize}
        
        \item \textbf{Selection rationale}: The \texttt{description} field in ChEBI-20-MM is written by chemistry experts and contains realistic molecular descriptions from scientific contexts (e.g., ``epoxy(hydroxy)icosatrienoate''), which helps avoid the templated and mechanical characteristics of manually constructed instructions. Meanwhile, the TOMG-Bench training data are highly aligned with the downstream evaluation tasks. By varying the mixing ratio, this experiment aims to identify a balance between realistic natural-language instructions and task-specific supervision, thereby reducing the risk of overfitting to a single data source and improving generalization.
    \end{itemize}

    \textbf{[Core Training Strategy]}
    \begin{itemize}
        \item Base model: \texttt{/<workspace>/0122205851-exp\_1/models/last} (the fully fine-tuned model obtained in Experiment 2)
        \item finetuning\_type: \texttt{full}
        \item learning\_rate: \texttt{2e-5}
        \item num\_gpus: \texttt{4}
        \item per\_device\_train\_batch\_size: \texttt{1}
        \item gradient\_accumulation\_steps: \texttt{16}
        \item num\_train\_epochs: \texttt{2}
        \item cutoff\_len: \texttt{2048}
        \item warmup\_ratio: \texttt{0.03}
        \item max\_grad\_norm: \texttt{1.0}
        \item work\_dir: \texttt{<workspace>}
        \item Other parameters are fixed across all runs, while the dataset is varied according to the following experimental settings:
        \begin{itemize}
            \item \textbf{Data Mixing Ratio Experiments (4 groups)}:
            \begin{itemize}
                \item \texttt{<workspace>/datasets/chebi\_100.jsonl}
                \item \texttt{<workspace>/datasets/chebi70\_tomg30.jsonl}
                \item \texttt{<workspace>/datasets/chebi40\_tomg60.jsonl}
                \item \texttt{<workspace>/datasets/chebi20\_tomg80.jsonl}
            \end{itemize}
        \end{itemize}
        \item Total number of experiments: $4$ groups.
    \end{itemize}
\end{mybox}